\documentclass[aos,MSNbibl,seceqn,dvips]{arximspdf}
\usepackage{dcolumn}
\usepackage{graphicx}
\doi{10.1214/13-AOS1163} %kopijuoti is PTS
\volume{41}
\issue{5}
\pubyear{2013}
\firstpage{2537}
\lastpage{2571}

\makeatletter
\newcolumntype{d}[1]{D{.}{.}{#1}}

\newproclaim{remark}{Remark}
\def\cal{\mathcal}

\newcommand{\hZ}{\hat{Z}}
\newcommand{\eps}{\varepsilon}
\newcommand{\goto}{\rightarrow}
\newcommand{\margmax}{\operatorname{argmax}}
\newcommand{\bphi}{\bar{\Phi}}
\newcommand{\sgn}{\operatorname{sgn}}
\newcommand{\cf}{\bar{F}}
\newcommand{\tf}{\widetilde{F}}
\newcommand{\tZ}{\tilde{Z}}
\newcommand{\tw}{\widetilde{W}}
\newcommand{\cg}{{\cal G}}
\newcommand{\bpsi}{\bar{\Psi}}
\newtheorem{thmm}{Theorem}[section] %(If you want theorem numbered
\newtheorem{lemma}{Lemma}[section] %% with section number.
\newtheorem{cor}{Corollary}[section]

\newproclaim{definition}{Definition}[section]
\makeatother

\begin{document}
\begin{frontmatter}

\title{Optimal classification in sparse Gaussian graphic model}
\runtitle{Optimal classification in sparse Gaussian graphic model}

\begin{aug}
\author[a]{\fnms{Yingying} \snm{Fan}\corref{}\thanksref{t1}\ead[label=e1]{fanyingy@marshall.usc.edu}},
\author[b]{\fnms{Jiashun} \snm{Jin}\thanksref{t2}\ead[label=e2]{jiashun@stat.cmu.edu}}
\and
\author[c]{\fnms{Zhigang} \snm{Yao}\thanksref{t3}\ead[label=e3]{zhigang.yao@epfl.ch}}
\thankstext{t1}{Supported in part by NSF CAREER Award DMS-11-50318,
Grant DMS-09-06784 and USC Marshall summer research funding.}
\thankstext{t2}{Supported in part by NSF Grant DMS-12-08315.}
\thankstext{t3}{Supported in part by NSF Award SES-1061387.}
\runauthor{Y. Fan, J. Jin and Z. Yao}
\affiliation{University of Southern California, Carnegie Mellon
University and Ecole~Polytechnique F\'{e}d\'{e}rale de Lausanne}
\address[a]{Y. Fan\\
Department of Data Sciences and Operations\\
Marshall School of Business\\
University of Southern California\\
Los Angeles, California 90089\\
USA\\
\printead{e1}}
\address[b]{J. Jin\\
Department of Statistics\\
Carnegie Mellon University\\
Pittsburgh, Pennsylvania 15213\\
USA\\
\printead{e2}}
\address[c]{Z. Yao\\
Section de Math\'{e}matiques\\
Ecole Polytechnique F\'{e}d\'{e}rale de Lausanne\\
EPFL Station 8, 1015 Lausanne\\
Switzerland \\
\printead{e3}}
\end{aug}

% HISTORY:
\received{\smonth{1} \syear{2013}}
\revised{\smonth{8} \syear{2013}}

% ABSTRACT

\begin{abstract}
Consider a two-class classification problem where the number of
features is much larger than the sample size.
The features are masked by Gaussian noise with mean zero and covariance
matrix $\Sigma$, where the precision matrix
$\Omega= \Sigma^{-1}$ is unknown but is presumably sparse. The useful
features, also unknown, are sparse and each contributes weakly (i.e.,
rare and weak) to the classification decision.

By obtaining a reasonably good estimate of $\Omega$, we formulate the
setting as a linear regression model. We propose a two-stage
classification method where we first select features
by the method of \textit{Innovated Thresholding} (IT), and then use the
retained features and Fisher's LDA for classification.
In this approach, a crucial problem is how to set the threshold of IT.
We approach this problem by adapting the recent innovation of Higher
Criticism Thresholding (HCT).

We find that when useful features are rare and weak, the limiting
behavior of HCT is essentially just as good as the limiting behavior of
ideal threshold, the threshold one would choose if the underlying
distribution of the signals is
known (if only). Somewhat surprisingly, when $\Omega$ is sufficiently
sparse, its off-diagonal coordinates usually do not have a major
influence over the classification decision.

Compared to recent work in the case where $\Omega$ is the identity
matrix [\textit{Proc. Natl. Acad. Sci. USA} \textbf{105} (2008)
14790--14795;
\textit{Philos. Trans. R. Soc. Lond. Ser. A Math. Phys. Eng. Sci.}
\textbf{367} (2009) 4449--4470], the current setting is much
more general, which needs a new approach and much more sophisticated analysis.
One key component of the analysis is the intimate relationship between
HCT and Fisher's separation. Another
key component is the tight large-deviation bounds for empirical
processes for data with unconventional correlation structures, where
graph theory on vertex coloring plays an important role.\vspace*{-2pt}
\end{abstract}
%

% KEYWORDS

\end{frontmatter}

%%%%%%%%%%%%%%
%s1 #&#
\section{Introduction}
\label{secIntro}
Consider a two-class classification problem, where we have~$n$ labeled
training samples $(X_i, Y_i), 1 \leq i \leq n$. Here, $X_i$ are
$p$-dimensional feature vectors and $Y_i \in\{-1, 1\} $ are the
corresponding class labels. For simplicity, we assume two classes are
\textit{equally likely}, and the data are centered so that
%%%%%%%%%
%%%%%%%%%
%%%%%%%%%
%%%%%%%%%
%
%e1.1 #&#
%
\begin{equation}
\label{model} X_i \sim N(Y_i \cdot\mu,
\Sigma_{p,p}),
\end{equation}
where $\mu$ is the contrast mean vector between two classes, and
$\Sigma
_{p,p}$ is the $p \times p$ covariance matrix.
Given a fresh feature vector
%%%%%%%%%%%%
%%%%%%%%%%%%
%%%%%%%%%%%%
%
%e1.2 #&#
%
\begin{equation}
\label{testsample} X \sim N(Y \cdot\mu, \Sigma_{p,p}),
\end{equation}
the goal is to train $(X_i, Y_i)$ to decide whether $Y = -1$ or $Y = 1$.
We denote $\Sigma_{p, p}^{-1}$ by $\Omega_{p, p}$, and whenever there
is no confusion, we drop the subscripts ``$p,p$'' (and also that of any
estimator of it, say, $\hat{\Omega}_{p,p}$).

We are primarily interested in the so-called ``$p \gg n$'' regime.
In many applications where $p \gg n$ (e.g., genomics), we observe the
following aspects.
\begin{itemize}
\item\textit{Signals are rare}. Due to large $p$, the useful features
(i.e., the nonzero coordinates of $\mu$) are rare. For example, for a
given type of cancer or disease, there are usually only a small number
of relevant features (i.e., genes or proteins). When we measure
increasingly more features, we tend to include increasingly more \textit{irrelevant} ones.
\item\textit{Signals are individually weak}. The training data can be
summarized by the $z$-vector
%
%e1.3 #&#
%
\begin{equation}
\label{DefineZ} Z = \frac{1}{\sqrt{n}} \sum_{i = 1}^n
Y_i X_i \sim N(\sqrt{n} \mu, \Sigma).
\end{equation}
Due to the small $n$, signals are weak in the sense that, individually,
the nonzero coordinates of $\sqrt{n} \mu$ are small or moderately large
at most.
\item\textit{Precision matrix $\Omega$ is sparse}.
Take Genetic Regulatory Network (GRN) for example. The feature vector
$X = (X(1), \ldots, X(p))'$ represents the expression levels of $p$
different genes, and is approximately distributed as $N(\mu, \Sigma)$.
For any $1 \leq i \leq p$, it is believed that for all except a few
$j$, $1 \leq j \leq p$, the gene pair $(i, j)$
are conditionally independent given all other genes. In other words,
each row of $\Omega$ has only a few nonzero entries, so $\Omega$ is
sparse \cite{Dempster1972}.
\end{itemize}
In many applications, $\Omega$ is unknown and has to be estimated. In
many other applications such as complicate disease or cancer, decades
of biomedical studies have accumulated huge databases which are
sometimes referred to as ``data-for-data''~\cite{LiLi2008}. Such
databases can be used to accurately estimate $\Omega$
independently of the data at hand, so $\Omega$ can be
assumed as known. In this paper, we investigate both the case where
$\Omega$ is known and the case where $\Omega$ is unknown. In either case,
we assume $\Omega$ has unit diagonals:
%
%e1.4 #&#
%
\begin{equation}
\label{Omegediag} \Omega(i,i) = 1, \qquad 1 \leq i \leq p.
\end{equation}
Such an assumption is only for simplicity, and we do not use such
information for inference.

%%%%%%%%%%%%%%%
%%%%%%%%%%%%%%%
%%%%%%%%%%%%%%%
%s1.1 #&#
\subsection{Fisher's LDA and modern challenges}
Fisher's Linear Discriminant Analysis
(LDA) \cite{Fisher1936} is a well-known method for classification,
which utilizes a weighted average of the test features
$L(X) = \sum_{j = 1}^p w(j) X(j)$,
and predicts $Y = \pm1$ if $L(X) > < 0$. Here, $w = (w(1), \ldots,
w(p))'$ is a preselected weight vector. Fisher showed that the optimal
weight vector satisfies
%
%e1.5 #&#
%
\begin{equation}
\label{Fisherw} w \propto\Omega\mu.
\end{equation}
In the classical setting where $n \gg p$,
$\mu$ and $\Omega$ can be conveniently estimated and Fisher's LDA is
approachable.
Unfortunately, in the modern regime where $p \gg n$, Fisher's LDA faces
immediate challenges.
\begin{itemize}
\item It is challenging to estimate $\Omega$ simply because there are
$O(p^2)$ unknown parameters but we have only $O(np)$ different measurements.
\item Even in the simplest case where $\Omega= I_p$, challenges
remain, as the signals are rare and weak. See \cite{DonohoJin2008} for
the delicacy of the problem.
\end{itemize}

The paper is largely focused on addressing the second challenge. It
shows that
successful classification can be achieved by simultaneously exploiting
the sparsity of $\mu$ (aka. signal sparsity) and the sparsity of
$\Omega
$ (aka. graph sparsity).
For the first challenge,
encouraging progresses have been made recently (e.g., \cite
{FriedmanHastieTibshirani2007,CaiLuo2011}), and the problem is more or
less settled. Still, the paper has a two-fold contribution along this
line. First, we show that the performances of the methods in \cite
{FriedmanHastieTibshirani2007,CaiLuo2011} can be substantially
improved if we add an additional re-fitting step; see details in
Section~\ref{secSimul}. Second, we carefully analyze how the errors in
estimating $\Omega$ may affect the classification results.

%%%%%%%%%%%%%%%
%%%%%%%%%%%%%%%
%%%%%%%%%%%%%%%
%%%%%%%%%%%%%%%
%s1.2 #&#
\subsection{Innovated thresholding}
\label{subsecITadd}
We wish to adapt Fisher's LDA to the current setting. Recall that the
optimal choice of weight vector is
$w \propto\Omega\mu$. If we have a reasonably good estimate of
$\Omega
$ (see Section~\ref{subsecOmega2} for more discussion on estimating~$\Omega$), say, $\hat{\Omega}$, all we need is a good estimate of
$\mu$.

When $\mu$ is sparse, one usually estimates it with some type of
thresholding \cite{DonohoJohnstone1994}. Let $Z$ be the training
$z$-vector as in (\ref{DefineZ}). For some threshold $t$ to be
determined, there are three obvious approaches to thresholding:
\begin{itemize}
\item\textit{Brute-force Thresholding} (BT). We apply thresholding to $Z$
directly using the so-called clipping rule \cite{DonohoJin2008}: $\hat
{\mu}_t^Z(i) = \sgn(Z(i)) 1\{|Z(i)| \geq t\}$. Alternatively, one may
use soft thresholding or hard thresholding.
However, numeric studies (e.g.,~\cite{DonohoJin2008}) suggest that
different thresholding schemes only have small differences in
classification errors,
provided that these schemes use the same threshold picked from
the range of interest.
For this reason, we only study the clipping rule; same below.
\item\textit{Whitened Thresholding} (WT). We first whiten the noise by
the transformation $Z \mapsto\hat{\Omega}^{1/2}Z \approx N(\sqrt{n}
\Omega^{1/2} \mu, I_p)$, and then apply the thresholding to the vector
$\hat{\Omega}^{1/2} Z$ in a similar fashion.
\item\textit{Innovated Thresholding} (IT). We first take the
transformation $Z \mapsto\hat{\Omega} Z$ and then apply the
thresholding by
%
%e1.6 #&#
%
\begin{equation}
\label{ITthreshold} \hat{\mu}_t^{\hat{Z}}(i) = \sgn \bigl(
\hat{Z}(i) \bigr) \cdot1\bigl\{\bigl | \hat {Z}(i)\bigr| \geq t \bigr\},\qquad \mbox{where }
\hat{Z} \equiv\hat{\Omega} Z.
\end{equation}
\end{itemize}
The transformation $Z \mapsto\hat{\Omega}Z$ is connected to the term
of \textit{Innovation} in the literature of time series \cite
{HallJin2010}, and so the name of Innovated Thresholding.

It turns out that, among the three approaches, IT is the best.
To see the point, note that for any $p \times p$ nonsingular matrix
$M$, one could always estimate $\mu$ by applying the thresholding to $M
Z$ entry-wise (in BT, WT, and IT, $M = I_p, \Omega^{1/2}$, and $\Omega$
approximately). The deal is, what is the best $M$?

Toward this end, write $M= [m_1, m_2, \ldots, m_p]'$. For any $1 \leq
i \leq p$, it is seen that
$(M Z)(i) \sim N(\sqrt{n} m_i' \mu, m_i' \Sigma m_i)$.
Therefore, if we bet on $\mu(i) \neq0$, we should choose $m_i$ to
optimize the
Signal-to-Noise Ratio (SNR)
of $(MZ)(i)$. By the Cauchy--Schwarz inequality, the optimal $m_i$
satisfies that
$m_i \propto\Omega\mu$.
Writing $\Omega= [\omega_1, \omega_2, \ldots, \omega_p]$, it is
seen that
%
%e1.7 #&#
%
\begin{equation}
\label{omegaterms} \Omega\mu= \mu(i) \omega_i + \sum
_{k \neq i} \mu(k) \omega_k \equiv (I) + (\mathit{II}).
\end{equation}
When we bet on $\mu(i) \neq0$, $(I) \propto\omega_i$ which is
accessible to us.
However, $(\mathit{II})$ is a very noisy vector and is inaccessible to us,
estimating which is equally hard as estimating
$\mu$ itself.

In summary, if we bet on $\mu(i) \neq0$, then the ``best'' accessible
choice is $m_i \propto\omega_i$. As this holds for all $i$ and we do
not know where the signals are, the optimal choice for $M$ is $M =
\Omega$. This says that IT is not only the best among the three choices
above, but is also the best choice in more general situations.

The heuristics above are consolidated in Sections~\ref{subsecOmega1}--\ref{subsecBTWT}, where we show that IT based
classifiers
achieve the optimal phase diagram for classification, while BT or WT
based classifiers do not, even in very simple settings.

\begin{remark*}The advantage of IT over WT and BT can be illustrated
with the following example, which is further discussed later in Section~\ref{subsecBTWT} where
we compare the phase diagrams of IT, WT, and BT.
Suppose $\Omega$ is a block diagonal matrix where for $h \in(-1,1)$
and $1 \leq i, j \leq p$,
%
%e1.8 #&#
%
\begin{eqnarray}
\label{blockwiseadd} \Omega(i,j) &= &1 \{ i = j \} + h \cdot1\{ j - i = 1,
\mbox{$i$ is odd} \}
\nonumber
\\[-8pt]
\\[-8pt]
\nonumber
&&{}+ h \cdot1\{ i-j = 1, \mbox{$i$ is even} \}.
\end{eqnarray}
According to the block structure of $\Omega$, we also partition the
vector $\mu$ into $p/2$ blocks, and each block has two entries. For
simplicity, we suppose each block of $\mu$ has either no signal, or a
single signal with a strength $\tau/\sqrt{n} > 0$.
BT, WT, and IT apply thresholding to $Z$, $\Omega^{1/2} Z$, and
$\Omega
Z$, correspondingly, where $Z \sim N(\sqrt{n} \mu, \Sigma)$ is the
training $z$-vector as above. In this simple example, the SNR
for $Z$, $\Omega^{1/2} Z$, and $\Omega Z$ are $\sqrt{(1 - h^2)} \tau$,
$[\sqrt{(1+h)} + \sqrt{(1 - h)}] \tau/2$ and
$\tau$ correspondingly, with the last one being the largest (for the
mean vector of $\Omega^{1/2} Z$ or $\Omega Z$, the nonzero coordinates
have two different magnitudes; the SNR is computed based on the larger
magnitude).
\end{remark*}

\begin{remark*} In (\ref{omegaterms}), the point that (II) is generally
noninformative in designing the best $m_i$ can be further elaborated
as follows: since we do not know
the locations of other nonzero coordinates of $\mu$, it makes sense to
model $\{ \sqrt{n} \mu(j)\dvtx 1 \leq j \leq p, j \neq i\}$ as i.i.d.
samples from
%
%e1.9 #&#
%
\begin{equation}
\label{eqmixture} (1 - \eps_p) \nu_0 +
\eps_p H_p, \qquad\mbox{$\eps_p > 0$: small},
\end{equation}
where $\nu_0$ is the point mass at $0$ and $H_p$ is some distribution
with no mass at $0$. Under general ``rare and weak'' conditions for $\mu
$ and sparsity condition for $\Omega$, entries of $E[(\mathit{II})]$ are
uniformly small.
\end{remark*}

In the literature of variable selection, IT is also called \textit{marginal regression}~\cite{Genovese2012}.
The connection is not surprising, as approximately, $\hat{\Omega}^{1/2}
Z \approx\Omega^{1/2} Z \sim N( \sqrt{n} \Omega^{1/2} \mu, I_p)$
which is a regression model.
Both methods apply thresholding to $\Omega Z$ entry-wise, but marginal
regression uses the hard thresholding rule, and IT uses the clipping
thresholding rule \cite{DonohoJin2008}.

With that being said, challenges remain on how to set the threshold $t$
of IT [see~(\ref{ITthreshold})]. If we set $t$ too small or too large,
the resultant estimator $\hat{\mu}_t^{\hat{Z}}$ has too many or too few
nonzeros. Our proposal is to set the threshold in a data driven fashion
by using the recent
innovation of Higher Criticism Thresholding (HCT).

%%%%%%%%%%
%%%%%%%%%%
%%%%%%%%%%
%%%%%%%%%%
%s1.3 #&#
\subsection{Threshold choice by higher criticism}
Higher Criticism (HC) is a notion mentioned in passing by Tukey \cite
{Tukey1976}. In recent years,
HC was found to be useful in sparse signal detection \cite
{DonohoJin2004}, large-scale multiple testing \cite{Ery,CaiJinLow2007,Zhong}, goodness-of-fit~\cite{JagerWellner2007},
and was applied to nonGaussian detection in Cosmic Microwave Background
\cite{Cayon2005} and genomics \cite{Wu,Sabatti2008}.
HC as a method for threshold choice in feature selection was first
introduced in
\cite{DonohoJin2008} (see also \cite{HallPittelkowGhosh2008}), but the
study has been focused on the case where $\Omega$
is the identity matrix. The case we consider in the current paper is
much more complicated, where how to use HC for threshold choice is a
nontrivial problem.

Our proposal is as follows. Let $\hat{\Omega}$ be a reasonably good
estimate of $\Omega$ and let $Z$ be the training $z$-vector as in
(\ref
{DefineZ}). As in (\ref{ITthreshold}), denote for short
%
%e1.10 #&#
%
\begin{equation}
\label{DefinehatZ} \hat{Z} = \hat{Z}(Z, \hat{\Omega}, p, n) = \hat{\Omega}Z.
\end{equation}
The proposed approach contains three simple steps.
\begin{itemize}
\item For each $1 \leq j \leq p$, obtain a $p$-value by $\pi_j =
P(|N(0,1)| \geq|\hat{Z}(j)|)$.
\item Sort all the $p$-values in the ascending order $\pi_{(1)} < \pi
_{(2)} < \cdots< \pi_{(p)}$.
\item Define the HC functional
$\operatorname{HC}_{p,j} = \sqrt{p} [j/p - \pi_{(j)} ]/ \sqrt{(1 - j/p) j / p}$, $1
\leq j \leq p$.
Let $\hat{j}$ be the index at which $\operatorname{HC}_{p,j}$ takes the maximum.
The Higher Criticism Threshold (HCT)---denoted by $|\hZ_{(\hat
j)}|$---is defined as
the $\hat{j}$th largest coordinate of $(|\hat{Z}(1)|, \ldots, |\hat
{Z}(p)|)'$.
\end{itemize}
Moreover, for stability, we need the following refinement. Define
%
%e1.11 #&#
%
\begin{equation}
\label{Definesnp} s_p^* = \sqrt{2 \log(p)},\qquad \tilde{s}_{p,n}^*
= \sqrt{2 \max\bigl\{0, \log\bigl(p / n^2\bigr) \bigr\}}.
\end{equation}
It is well-understood (e.g., \cite{DonohoJin2004,HallJin2010}) that
the threshold should not be larger than~$s_p^*$. At the same time, the
threshold should not be too small, especially when $n$ is small.
The HCT we use in this paper is
%
%e1.12 #&#
%
\begin{equation}
\label{HCTrefined} t_p^{\mathrm{HC}} = \cases{|\hZ_{(\hat j)}|, &\quad  $\mbox{if } \tilde{s}_{p,n}^*
\leq|\hZ_{(\hat
j)}| \leq s_p^*$,
\vspace*{2pt}\cr
\tilde{s}_{p,n}^*, & \quad$\mbox{if } |\hZ_{(\hat j)}| <
\tilde{s}_{p,n}^*$,
\vspace*{2pt}\cr
s_p^*, &\quad  $\mbox{if } |\hZ_{(\hat j)}| > s_p^*.$}
\end{equation}
See Sections~\ref{subsecmodel} and~\ref{secHCT} for more
detailed discussion.

%%%%%%%%%%%%%%
%%%%%%%%%%%%%%
%%%%%%%%%%%%%%
%%%%%%%%%%%%%%
%%%%%%%%%%%%%%
%s1.4 #&#
\subsection{HCT trained classifier}
We are now ready for classification. Let $\hat{\Omega}$ be as above,
and let $\hat{\mu}_{\mathrm{HC}}^{\hat{Z}} = \hat{\mu}^{\hat{Z}}(Z, \hat
{\Omega
}, p, n)$ be defined as
%
%e1.13 #&#
%
\begin{equation}
\label{hctclassadd1} \hat{\mu}_{\mathrm{HC}}^{\hat{Z}}(j) = \sgn\bigl(
\hat{Z}(j)\bigr) \cdot1\bigl\{ \bigl|\hat {Z}(j) \bigr| \geq t_p^{\mathrm{HC}}
\bigr\},\qquad 1 \leq j \leq p.
\end{equation}
Compared to $\hat{\mu}_t^{\hat{Z}}$ in (\ref{ITthreshold}), the only
difference is that we have replaced $t$ by $t_p^{\mathrm{HC}}$.
Introduce the HCT classification statistic
%
%e1.14 #&#
%
\begin{equation}
\label{hctclassadd2} L_{\mathrm{HC}}(X, \hat{\Omega}) = L_{\mathrm{HC}}(X,
\hat{\Omega}; Z, p, n) = \bigl(\hat {\mu }_{\mathrm{HC}}^{\hat{Z}}
\bigr)' \hat{\Omega} X.
\end{equation}
The HCT trained classifier (or HCT classifier for short) is then the
decision rule that decides $Y = \pm1$ according to $L_{\mathrm{HC}}(X, \hat
{\Omega}) > < 0$.

The innovation of the procedure is two-fold: using IT for feature
selection and using HCT for threshold choice in the more complicated
case where $\Omega$ is unknown and is nonidentity. The work is
connected to other works on HC \cite{HallJin2010,DonohoJin2008}, but
the procedure and the delicate theory it entails are new.

A question is whether IT has any advantages over exsiting variable
selection methods (e.g., the Lasso \cite{Tibshirani1996}, SCAD \cite
{fan2}, Dantzig selector \cite{candes1}). The answer is yes, for the
following reasons. First, compared to these methods, IT is
computationally much faster and much more approachable for delicate
analysis. Second, our goal is classification, not variable selection.
For classification, especially when features are rare and weak, the
choice of different variable selection methods is secondary, while the
choice of the tuning parameter is crucial. The threshold of IT can be
conveniently set by HCT, but
how to set the tuning parameter of the Lasso, SCAD, or Dantzig selector
remains an open problem, at least in theory.

How does the HCT classifier behave? In Sections~\ref{subsecmodel}--\ref
{subsecLB}, we set up a theoretic framework and derive a lower bound
for classification errors. In Sections~\ref{subsecOmega1}--\ref{subsecOmega2}, we investigate the HCT
classifier for the cases where $\Omega$ is known and unknown
separately, and show that the HCT classifier yields optimal phase
diagram in classification.

%%%%%%%%%%%%%
%%%%%%%%%%%%%
%%%%%%%%%%%%%
%%%%%%%%%%%%%
%s1.5 #&#
\subsection{Asymptotic rare and weak model} \label{subsecmodel}
Motivated by the application examples aforementioned, we use a \textit{Rare and Weak} signal model as follows.
We model the scaled contrast mean vector $\sqrt{n} \mu$ as
%
%e1.15 #&#
%
\begin{equation}
\label{DefineH} \sqrt{n} \mu(j) \stackrel{\mathrm{i.i.d.}} {\sim} (1 -
\eps_p) \nu_0 + \eps_p H_p,\qquad 1
\leq j \leq p,
\end{equation}
where as in (\ref{eqmixture}), $\nu_0$ is the point mass at $0$, $H_p$
is some distribution with no mass at~$0$, and $\eps_p \in(0,1)$ is
small [note that $(\eps_p, H_p)$ depend on $p$ but not on $j$]. We use
$p$ as the driving asymptotic parameter, and link parameters $(n, \eps
_p, H_p)$ to $p$ through some fixed parameters.
In detail,
fixing parameters $(\beta, \theta) \in(0,1)^2$, we model
%
%e1.16 #&#
%
\begin{equation}
\label{Defineepsandn} \eps_p = p^{-\beta},\qquad n = n_p
= p^{\theta}.
\end{equation}
As $p$ tends to $\infty$, the sample size $n_p$ grows to $\infty$ but
in a slower rate than that of~$p$; the signals get increasingly sparser but the number of signals
tends to $\infty$.
The interesting range of parameters $(\beta, \theta, H_p)$ partitions
into three regimes, according to the sparsity level.
\begin{itemize}
\item\textit{Relatively Dense} (RD). In this regime, $0 < \beta< (1 -
\theta)/2$. The signals are relatively dense
and successful classification is possible even when signals are very
faint [e.g., $H_p$ concentrates its mass around a term $\tau_p \ll
\sqrt {2 \log(p)}$]. In such cases, (a) successful feature selection is
impossible as
signals are too weak, and (b) feature selection is unnecessary for the
signals are relatively dense.
\item\textit{Rare and Weak} (RW). In this regime, $(1 - \theta)/2 <
\beta
< (1 - \theta)$, and the signals are moderately sparse. For successful
classification, we need moderately strong signals [i.e., nonzero
coordinates of $\sqrt{n} \mu\asymp\sqrt{\log(p)}$]. In this case,
feature selection is subtle but could be substantially helpful.
In contrast, classification is impossible if signals are much weaker
than $\sqrt{\log(p)}$, and consistent feature selection is possible
(and so the problem of classification is much less challenging) if the
signals are much stronger than $\sqrt{\log(p)}$.
\item\textit{Rare and Strong} (RS). In this regime, $\beta> (1 -
\theta
)$, and the signals are very sparse. For successful classification, we
need very strong signals [signal strength $\gg\sqrt{\log(p)}$]. In
this case, feature selection is
comparably easier to carry out (but substantially helpful) since the
signals are strong enough to stand out for themselves.
\end{itemize}
While the statements hold broadly, the most transparent way to
understand them is probably to consider the case where $H_p$ is a point
mass at $\tau_p$ (say): in the above three regimes, the minimum $\tau
_p$ required for successful classification (up to some multi-$\log(p)$
factors in the first and last regimes) are
$1/(\eps_p \sqrt{(p/n_p)})$, $\sqrt{\log(p)}$, and $\sqrt{n_p /(p
\eps
_p)}$ correspondingly;
the proof is elementary so is omitted.

In summary, feature selection is impossible in the RD regime and is
relatively easy in the RS regime. For these reasons, we are primarily
interested in the RW regime where we assume
%
%e1.17 #&#
%
\begin{equation}
\label{varthetarange} (1 - \theta)/2 < \beta< (1 - \theta).
\end{equation}
The RD/RS regimes are further discussed in Section~\ref{subsecDiscu},
where we address the connection between our work and \cite{CaiLiu2011,FanFengTong2012,ShaoWangDengWang11}.
For $\beta$ in this range, the most interesting range for the signal strength
is when $H_p$ concentrates its mass at the scale of $\sqrt{\log(p)}$.
In light of this, we fix $r > 0$ and calibrate the signal strength
parameter~$\tau_p$ by
%
%e1.18 #&#
%
\begin{equation}
\label{Definetau} \tau_p = \sqrt{2 r \log(p)}.
\end{equation}

Except in Section~\ref{subsecLB} where we address the lower bound
arguments, we assume~$H_p$ is a point mass [compare (\ref{DefineH})]:
%
%e1.19 #&#
%
\begin{equation}
\label{pointmass}\qquad H_p = \nu_{\tau_p}, \qquad\mbox{where as in (
\ref{Definetau}), $\tau _p = \sqrt{2 r \log(p)}$ and $0 < r < 1$}.
\end{equation}
We focus on the case $0 < r < 1$, as the case $r > 1$ corresponds to
RS regime where the classification is comparably easier.
This models a setting where
the signal strengths are equal. The case where the signal strengths are unequal
is discussed in Section~\ref{subsecSummary}.

Next, we model $\Omega$. Motivated by the previous example on Genetic
Regulatory Network,
we assume each row of $\Omega$ has relatively few nonzeros. Such a
matrix naturally induces a sparse graph $\cg= \cg(\Omega) = (V, E)$,
where $V = \{1, 2, \ldots, p\}$ and there is an edge between nodes $i$
and $j$ if and only if $\Omega(i,j) \neq0$; see \cite{Bollobas}
for basic terminology in graph theory.

%de1.1 #&#
\begin{definition}\label{de1.1}
Fix $p$ and $1 \leq K < p$. We call a $p \times p$ positive definite
matrix $\Omega$ $K$-sparse if each row of $\Omega$ has at most $K$
nonzeros. For any graph $\cg$, we call $\cg$ $K$-sparse if the degree
of each node $\leq K$.
\end{definition}
When $\Omega$ is $K$-sparse, the induced graph $\cg(\Omega)$ is $(K-1)$
sparse, since by convention, there is no edge between a node and itself.

The class of $K$-sparse graphs is much broader than the class of banded
graphs (we call $\cg$ a banded graph with bandwidth $K$ if nodes $i$
and $j$ are not connected whenever $|i - j | > K$). In fact, even when
$\cg$ is $K$-sparse with $K = 2$, we cannot always shuffle the nodes
of $\cg$ and make it a banded graph with a small bandwidth.

Let ${\cal M}_p$ be the class of all $p \times p$ positive definite
correlation matrices. Fixing $a \in(0,1)$, $b > 0$, and
a sequence of integers $K_p$, introduce
%
%e1.20 #&#
%
\begin{equation}
\label{eqdefMset0} {\cal M}_p^*(a, K_p) = \bigl\{
\mbox{$\Omega\in{\cal M}_p$ and is $K_p$-sparse}, \bigl|
\Omega(i,j)\bigr| \leq a, i \neq j\bigr\}
\end{equation}
and
%
%e1.21 #&#
%
\begin{equation}
\label{eqdefMset} {\widetilde{\cal M}}_p^*(a, b, K_p) =
\bigl\{\Omega\in{\cal M}_p^*(a,K_p), \bigl\|
\Omega^{-1} \bigr\| \leq b \bigr\},
\end{equation}
where $\|\cdot\|$ is the spectral norm. In comparison, ${\widetilde
{\cal M}}_p^*(a, b, K_p)$ is slightly smaller than ${\cal M}_p^*(a, K_p)$.
The following
short-hand notation is frequently used in this paper.
%
%de1.2 #&#
\begin{definition}\label{de1.2}
We use $L_p$ to denote a strictly positive generic multi-$\log(p)$ term
that may vary from occurrence to occurrence but always satisfies that
for any fixed $c > 0$, $\lim_{p \goto\infty} \{ L_p p^{-c} \} = 0$ and
$\lim_{p \goto\infty} \{ L_p p^{c} \} = \infty$.
\end{definition}
In this paper, we are primarily interested in the case where $K_p$ is
at most multi-logarithmically large unless stated otherwise:
%
%e1.22 #&#
%
\begin{equation}
\label{DefineK} \lim_{p \goto\infty} K_p = \infty,\qquad
K_p \leq L_p;
\end{equation}
the first requirement is only for convenience.
In our classification setting, $X_i \sim N(Y_i \mu, \Sigma)$, $X \sim
N(Y\mu, \Sigma)$, and $Y = \pm1$ with equal probabilities. The
following notation is frequently used in the paper.
%
%de1.3 #&#
\begin{definition}\label{de1.3}
We say the classification problem (\ref{model})--(\ref{testsample})
satisfies the Asymptotic Rare Weak model $\operatorname{ARW}(\beta, r, \theta,
\Omega
)$ if (\ref{DefineH})--(\ref{Defineepsandn}), (\ref{pointmass}) and
(\ref{DefineK}) hold.
\end{definition}
\begin{remark*} The normalization in ARW is different from that in
conventional asymptotic settings. In the latter, we usually fix $\mu$
and let $n$ increase, so the classification problem becomes
increasingly easier as $n$ increase. In ARW, to focus on the ``most
interesting parameter regime'', we fix
$\sqrt{n} \mu$ and let $n$ increase. As a result, the SNR in the
summarizing training $Z$-vector remain the same, but the SNR in the
testing vector $X$ decrease rapidly as $n$ increase. Therefore, the
classification problem becomes increasingly harder as $n$ increase.
\end{remark*}
%%%%%%%%%%%
%%%%%%%%%%%
%%%%%%%%%%%
%s1.6 #&#
\subsection{Lower bound} \label{subsecLB}
Introduce the \textit{standard phase boundary} function
%
%e1.23 #&#
%
\begin{equation}
\label{eqstdbdry} \rho(\beta) = \cases{ %
 0, &\quad $0
< \beta\leq1/2$,
\vspace*{2pt}\cr
\beta- 1/2, &\quad $1/2 < \beta< 3/4$,
\vspace*{2pt}\cr
(1 - \sqrt{1 - \beta})^2, &\quad $3/4 \leq\beta< 1,$}
\end{equation}
and let
\[
\rho_{\theta}^{*}(\beta) = (1 - \theta) \rho\bigl(\beta/(1 -
\theta)\bigr), \qquad (1 - \theta) / 2 < \beta< (1 - \theta).
\]
The function $\rho$ has appeared before in determining
phase boundaries in a seemingly unrelated
problem of multiple hypothesis testing \cite{Ingster1997,Ingster1999,DonohoJin2004}.
The following theorem is proved in the supplementary material \cite{HCTghsupp}.
%%%%%%%%%%%%%
%%%%%%%%%%%%%
%%%%%%%%%%%%%
%%%%%%%%%%%%%
%
%th1.1 #&#
\begin{thmm} \label{thmmLB}
Fix $(\beta, r, \theta) \in(0,1)^3$ such that $(1 - \theta)/2 <
\beta
< (1 - \theta)$ and
$0 < r < \rho_{\theta}^*(\beta)$. Suppose (\ref{DefineH})--(\ref
{Defineepsandn}), (\ref{Definetau}), and
(\ref{DefineK}) hold and that for sufficiently large $p$, $\Omega\in
{\cal M}_p^*(a, K_p)$ and the support of $H_p$ is contained in $[-\tau
_p, \tau_p]$. Then as $p \goto\infty$, for any sequence of trained
classifiers, the misclassification error $\gtrsim1/2$.
\end{thmm}
Note that in Theorem~\ref{thmmLB}, we do not require the signals to
have the same strength.
Also, recall that in our classification setting (\ref{model})--(\ref
{testsample}), two classes are assumed as equally likely; extension to
the case where two classes are unequally likely is straightforward.
Theorem~\ref{thmmLB} was discovered before in \cite{DonohoJin2008,ingster2009b}, but the study has been focused on the case where $\Omega
= I_p$ and $H_p$ is the point mass at $\tau_p$. The proof in the
current case is much more difficult and needs a few tricks, where graph
theory on vertex coloring plays a key role. The following lemma is
adapted from \cite{Bollobas}, Section V.1.
%%%%%%%%%%%
%%%%%%%%%%%
%%%%%%%%%%%
%
%le1.1 #&#
\begin{lemma} \label{lemmapartition}
Fix $K \geq1$. For any graph $\cg= (V, E)$ that is $K$-sparse, the
chromatic number of $\cg$ is no greater than $(K+1)$.
\end{lemma}
Recall that when $\Omega$ is $K$ sparse, then the induced graph $\cg=
\cg(\Omega)$ is
$(K - 1)$ sparse, and so the chromatic number of $\cg(\Omega)$ $\leq K$.
As a result, we can color the nodes of $\cg(\Omega)$ with no more than
$K$ different colors,
where there is no edge between any pair of nodes with the same color.

Despite its seemingly simplicity,
Lemma~\ref{lemmapartition} has far-reaching implications. Lemma~\ref
{lemmapartition} is the corner stone for proving
the lower bound and for analyzing the HCT classifier (where we need
tight convergence rate of empirical processes for data with
unconventional correlation structures).

%%%%%%%%%%%%%%%%
%%%%%%%%%%%%%%%%
%%%%%%%%%%%%%%%%
%%%%%%%%%%%%%%%%
%s1.7 #&#
\subsection{\texorpdfstring{HCT achieves optimal phase diagram in classification
($\Omega$ is known)}{HCT achieves optimal phase diagram in classification
(Omega is known)}} \label{subsecOmega1}
One noteworthy aspect of HCT classifier is that it achieves the optimal
phase diagram. In this section, we show this for the case where
$\Omega$ is known. In this case, the HCT classifier $L_{\mathrm{HC}}(X, \hat
{\Omega})$ reduces to $L_{\mathrm{HC}}(X, \Omega)$ (the term formed by replacing
$\hat{\Omega}$ by $\Omega$ everywhere in the definition of former).
Since we predict the label associated with $X$ as $\pm1$ according to
$L_{\mathrm{HC}}(X, \Omega) > < 0$, the predicted label is correct if and only if
$Y \cdot L_{\mathrm{HC}}(X, \Omega) > 0$. The following theorem is proved in
Section~\ref{subsecclassify}.
%$L_{\mathrm{HC}}(X, \Omega)$ and the true label variable $Y$ take values from $
%{\mathrm{HC}}(X, \Omega)$ if and only if $Y \cdot L_{\mathrm{HC}}(X, \Omega) > 0$.
%%%%%%%%%%%%%%
%%%%%%%%%%%%%%
%%%%%%%%%%%%%%
%%%%%%%%%%%%%%
%
%th1.2 #&#
\begin{thmm} \label{thmmIdealized}
Fix $(\beta, r, \theta, a) \in(0,1)^4$ such that $(1 - \theta)/2 <
\beta< (1 - \theta)$ and $r > \rho_{\theta}^*(\beta)$. Consider a
sequence of classification problems $\operatorname{ARW}(\beta, r, \theta, \Omega)$ with
$\Omega\in\widetilde{\cal M}_p^*(a, b, K_p)$ for sufficiently large
$p$. Then as $p$ tends to $\infty$,
$P  ( Y \cdot L_{\mathrm{HC}}(X, \Omega) < 0  ) \goto0$.
When $r <\beta$, the condition on $\Omega$ can be relaxed to that of
$\Omega\in{\cal M}_p^*(a,K_p)$.
\end{thmm}

Call the two-dimensional space $\{(\beta, r)\dvtx 0 < \beta< 1, 0 < r <
1\}
$ the phase space.
Theorems \ref{thmmLB}--\ref{thmmIdealized} say that the phase space
partitions into two separate regions, \textit{Region of Impossibility} and
\textit{Region of Possibility}, where the classification problem is
distinctly different.
\begin{itemize}
\item\textit{Region of Impossibility}. $\{(\beta, r)\dvtx (1 - \theta
)/2 <
\beta< (1 - \theta), 0 < r < \rho_{\theta}^*(\beta)\}$.
Fix $(\beta, r)$ in the interior of this region and consider a sequence
of classification problems
with $p^{1 - \beta}$ signals where each signal $\leq\sqrt{2 r \log
(p)}$ in strength. Then for any sequence of ``sparse'' $\Omega$,
successful classification is impossible. This is the most difficult
case where not much can be done for classification aside from random guessing.
\item\textit{Region of Possibility}. $\{(\beta, r)\dvtx (1 - \theta)/2 <
\beta< (1 - \theta)\}, \rho_{\theta}^*(\beta) < r < 1\}$.
Fix $(\beta, r)$ in the interior of this region and suppose signals
have equal strength of $\sqrt{2 r \log(p)}$. HCT classifier $ L_{\mathrm{HC}}(X,
\Omega)$ yields successful classification (the results hold much more
broadly where equal signal strength assumption can be largely relaxed).
\end{itemize}
We call the curve $r = \rho_{\theta}^*(\beta)$ the \textit
{separating boundary}.
Somewhat surprisingly, the separating boundary does not depend on the
off-diagonals of $\Omega$.
The partition of phase diagram was discovered by \cite
{DonohoJin2008,JinPNAS2009}, and independently by \cite{ingster2009b},
but the focus was on the case where $\Omega= I_p$. See also \cite
{HallPittelkowGhosh2008}. The study in the current case is much more
difficult. Similar phase diagrams are also found in sparse signal
detection \cite{DonohoJin2004}, variable selection \cite{jijin2010},
and spectral clustering \cite{jinwang2012}.

Why HCT works? The key insight is that there is an intimate
relationship between the HC functional and Fisher's separation; the
latter plays a key role in determining the optimal classification
behavior, but is, unfortunately, an \textit{oracle} quantity which depends
on unknown parameters.
In Sections~\ref{secIdealHC}--\ref{secHCT}, we outline a series of
theoretic results, explaining why the HCT classifier is the right
approach and how it achieves the optimality.

%%%%%%%%%%%%%
%%%%%%%%%%%%%
%%%%%%%%%%%%%
%s1.8 #&#
\subsection{\texorpdfstring{Optimality of HCT classification ($\Omega$ is unknown)}
{Optimality of HCT classification (Omega is unknown)}}
\label{subsecOmega2}
%%%%%%%%%%%%%%%%
%%%%%%%%%%%%%%%%
%%%%%%%%%%%%%%%%
When $\Omega$ is unknown, we first estimate it with the training data.
%
%de1.4 #&#
\begin{definition}\label{de1.4}
For any sequence of $\Omega_{p,p} \in{\cal M}_p^*(a, K_p)$, we say an
estimator $\hat{\Omega}_{p,p}$ is acceptable if it is symmetric and
independent of the test feature vector~$X$, and that there is
a constant $C > 0$ such that for sufficiently large $p$,
$\hat{\Omega}_{p,p}$ is $K_p'$-sparse where $K_p' \leq L_p$, and
$|\hat{\Omega}_{p,p}(i,j) - \Omega_{p,p}(i,j)| \leq C K_p^2 \sqrt {\log
(p)}/\sqrt{n_p}$ for
all $1 \leq i, j \leq p$.
\end{definition}
Usually, the $(L_p/\sqrt{n_p})$-rate cannot be improved, even when
$\Omega$ is diagonal.
For $K_p$-sparse $\Omega$ satisfying (\ref{DefineK}), acceptable
estimators can be constructed based on the recent CLIME approach in
\cite{CaiLuo2011}.
If additionally $\Omega$ satisfies the mutual incoherence condition
\cite{RavikumarWainwrightRaskuttiYu2011}, Assumption~1,
then the glasso \cite{FriedmanHastieTibshirani2007} is also acceptable,
provided the tuning parameters are properly set.
If $\Omega$ is banded, then the
Bickel and Levina Thresholding (BLT) method \cite{Bickel2008} is also
acceptable, up to some modifications.

With that being said, the numeric performances of all these estimators
can be improved with an additional step of \textit{re-fitting}. See
Section~\ref{secSimul} for details.

Naturally, the estimation error of $\hat{\Omega}$ has some negative
effects on the HCT classifier. Fortunately, for a large fraction of
parameters $(\beta, r)$ in Region of Possibility, such
effects are negligible and HCT continues to yield successful classification.
In detail, recalling that $n_p = p^{\theta}$, $\theta\in(0, 1)$, we
suppose:
\begin{itemize}
\item\textit{Condition} (a). $r > \max\{ (1 - 2 \theta)/4, \rho
_{\theta
}^*(\beta) \}$,
\item\textit{Condition} (b). When $0 < \theta\leq1/3$ and $(1 -
\theta
)/2 < \beta< (1 - 2 \theta)$, $|r - \sqrt{1 - 2 \theta}| \geq\sqrt{1
- 2 \theta- \beta}$.
\end{itemize}
The following theorem is proved in Section~\ref{subsecclassify}.
%%%%%%%%%%%
%%%%%%%%%%%
%%%%%%%%%%%
%%%%%%%%%%%
%
%th1.3 #&#
\begin{thmm} \label{thmmUB1}
Fix $(\beta, r, \theta, a) \in(0,1)^4$ such that $(1 - \theta)/2 <
\beta< (1 - \theta)$, and conditions \textup{(a)}--\textup{(b)} hold. Consider a sequence
of classification problems $\operatorname{ARW}(\beta, r, \theta, \Omega)$ such that
$\Omega\in{\cal M}_p^*(a, K_p)$ when $r <\beta$ and
$\Omega\in\widetilde{\cal M}_p^*(a, b, K_p)$ when $r \geq\beta$. For
the HCT classifier $L_{\mathrm{HC}}(X, \hat{\Omega})$, if $\hat{\Omega}$ is
acceptable, then as $p$ tends to $\infty$, $P( Y \cdot L_{\mathrm{HC}}(X, \hat
{\Omega}) < 0 ) \goto0$.
\end{thmm}
We remark that, first, when $0 < \theta\leq1/4$ and $(1 - \theta)/2 <
\beta< 3(1 - 2 \theta)/4$, condition (a) can be relaxed to that of $r
> \max\{\beta/3, \rho_{\theta}^*(\beta)\}$. Second,
when $\theta> 1/2$, conditions (a)--(b) automatically hold when $r >
\rho_{\theta}^*(\beta)$. As a result, we have the following corollary,
the proof of which is omitted.
%
%co1.1 #&#
\begin{cor} \label{corUB}
When $\theta> 1/2$, Theorem~\ref{thmmUB1} holds with conditions
\textup{(a)}--\textup{(b)} replaced by that of $r > \rho_{\theta}^*(\beta)$.
\end{cor}
This says that as long as $n_p \gg\sqrt{p}$, the estimation errors of
any acceptable estimator $\hat{\Omega}$ have negligible effects over
the classification decision.

%%%%%%%%%%%%%%%
%%%%%%%%%%%%%%%
%%%%%%%%%%%%%%%
%%%%%%%%%%%%%%%
%s1.9 #&#
\subsection{Comparison with BT and WT}
\label{subsecBTWT}
In disguise, many methods are what we called ``Brute-forth Thresholding''
or ``BT,'' including but not limited to \cite{Bickel2004,Efron09,FanFan2008,TibshiraniPNAS2002}.
Since $\Omega$
is hard to estimate, Bickel and Levina \cite{Bickel2004}, Fan and Fan
\cite{FanFan2008}, and Tibshirani {et al.} \cite{TibshiraniPNAS2002}
neglect the off-diagonals in $\Sigma$ for classification. In a
seemingly different spirit, Efron \cite{Efron09} proposes
a procedure where he first selects features by neglecting the
off-diagonals in $\Sigma$
and then estimates the correlation structures among selected features.
However, under the Rare and Weak model,
selected features tend to be uncorrelated. Therefore, at least for many
cases, the approach fails to exploit the ``local'' graphic structure of
the data and is ``BT'' in disguise.
It is also noteworthy that \cite{TibshiraniPNAS2002} proposes to set
the threshold of BT by cross validation, which is unstable, especially
when $n_p$ is small.

When we replace IT by either BT or WT in HCT classifier, the phase
diagram associated with the resultant procedure is no longer optimal.
While the claim holds very broadly,
it can be conveniently illustrated with a simple example as follows.

Consider the same setting as in Theorem~\ref{thmmIdealized}, except that
$\Omega$ is the matrix defined in (\ref{blockwiseadd}). That is,
$\Omega$ is
the diagonal block-wise matrix where each diagonal block is the $2
\times2$ matrix
with $1$ on the diagonals and $h$ on the off-diagonals, $h \in(-1, 1)$.
In this simple case, by Theorem~\ref{thmmIdealized}, HCT classifier gives
successful classification when $r > \rho_{\theta}^*(\beta)$, and
fails when
$r < \rho_{\theta}^*(\beta)$.
In comparison, if we use BT (which treats $\Sigma$ as diagonal and
does not
incorporate correlations for classification), the separating function
for success
and failure becomes $r = \rho_{\theta}^*(\beta) /(1 - h^2)$, which is
higher than
$r = \rho_{\theta}^*(\beta)$ in the $\beta$-$r$ plane
(a similar claim holds for WT, but the separating function is $r = 2
\rho_{\theta}^*(\beta)/ [1 +\sqrt{1-h^2}]$; note $2/[1 + \sqrt{1 -
h^2}] >1$ for all $h \neq0$).
Recall that when $\Omega$ is given, the \textit{only} difference between
the HCT classifier built over IT and the HCT classifier built over BT
is that, for any threshold $t$, BT and IT estimate $\mu$ by
\[
\hat{\mu}_t^{Z}(i) = \sgn\bigl(Z(i)\bigr) 1\bigl\{\bigl|Z(i)\bigr|
\geq t\bigr\}\quad \mbox{and}\quad \hat{\mu}_t^{\tilde{Z}}(i) = \sgn\bigl(
\tilde{Z}(i)\bigr) 1\bigl\{ \bigl|\tilde{Z}(i)\bigr| \geq t\bigr\},
\]
respectively,
where $\tilde{Z} = \Omega Z$; see Section~\ref{subsecITadd} for details.
We have the following theorem, the proof of which is elementary so is omitted.
%%%%%%%%%
%%%%%%%%%
%%%%%%%%%
%
%th1.4 #&#
\begin{thmm} \label{thmmBickel}
Fix $(\beta, \theta, r) \in(0,1)^3$ such that $(1 - \theta)/2 <
\beta
< (1 - \theta)$. Consider a sequence of
classification problems $\operatorname{ARW}(\beta, r, \theta, \Omega)$ where
$\Omega$
is the diagonal block-wise matrix defined in (\ref{blockwiseadd}).
Suppose we apply HCT classifier built over the Brute-force Thresholding
(BT) as in Section~\ref{subsecITadd}.
As $p \goto\infty$, the classification error $\goto0$ if $r > \rho
_{\theta}^*(\beta)/(1 - h^2)$,
and the classification error $\goto1/2$ if $r < \rho_{\theta}^*/(1 - h^2)$.
\end{thmm}
%
%%%%%%%%%
%%%%%%%%%
%%%%%%%%%
%s1.10 #&#
\subsection{Comparison with works focused on the RS regime}
\label{subsecDiscu}
The work is closely related to the recent approach by Shao {et al}.
\cite{ShaoWangDengWang11},
the ROAD approach by Fan {et al}. \cite{FanFengTong2012},
and the LPD approach by Cai and Liu \cite{CaiLiu2011}.
While all approaches attempt to mimic Fisher's LDA,
the difference lies in how we estimate the ``ideal weight vector'' $w$
prescribed in (\ref{Fisherw}). In our notation,
Shao {et al}. \cite{ShaoWangDengWang11} estimates $w$ by
$(\Sigma
^*)^{-1} \hat{\mu}_t^{Z}$,
where $\Sigma^*$ is the regularized estimation of $\Sigma$ as in
Bickel and Levina \cite{Bickel2008} for an appropriate threshold, and
$\hat{\mu}_t^{Z}$ is the estimation of $\mu$ by Brute-force
Thresholding.
ROAD estimates $w$
by minimizing $(1/2) w' \hat{\Sigma} w + \lambda\| w\|_1 + (1/2)
\gamma(w' Z - 1)^2$,
and LPD estimates $w$ by minimizing $\|w\|_1$ subject to the constraint of
$\|\hat{\Sigma} \beta- Z\|_{\infty} \leq\lambda$, where $\lambda
$ and
$\gamma$ are tuning parameters.

In disguise, these works focused on the ``Rare and Strong'' regime
according to our terminology. In fact, Shao {et al}. \cite
{ShaoWangDengWang11}
assumes the minimum signal strength (smallest coordinate in magnitude
of $\sqrt{n_p} \mu$) is of the order of $\sqrt{n_p}$, and the main
results of Fan {et al}. \cite{FanFengTong2012} and
Cai and Liu \cite{CaiLiu2011} (i.e., \cite{FanFengTong2012}, Theorem~3, \cite{CaiLiu2011}, Theorem~1)
assume a sparsity constraint that can be roughly translated to $\beta>
(1 - \theta/2)$ in our
notation. Seemingly, this concerns the RS Regime we mentioned earlier.

Compared to these works, our work focuses on the most challenging
regime where the signals are Rare and Weak, and we need much more
sophisticated methods for feature selection and for threshold choices.

%s1.11 #&#
\subsection{Comparison with other popular classifiers}
HCT classifier also has advantages over other well-known classifiers
such as the Support Vector Machine (SVM) \cite{Burges1998}, Random
Forest \cite{Breiman2001}, and Boosting \cite{Dettling2003}. These
methods need tuning parameters and are internally very complicated,
but they do not outperform HCT classifier even when we replace the IT
by BT; see
details in \cite{DonohoJin2008}, where
we have compared all these methods with three well-known gene
microarray data sets in the context of cancer classification.

HCT is also closely related to PAM \cite{TibshiraniPNAS2002}, but is
different in some important aspects. First, HCT exploits the
correlation structure while PAM does not. Second, while both methods
perform feature selection, PAM sets the threshold by cross validations
(CVT), and HCT sets the threshold by Higher
Criticism. When $n$ is small, CVT is usually unstable. In \cite
{DonohoJin2008}, we have
shown that HCT outperforms CVT when analyzing three microarray data
sets aforementioned.
In Section~\ref{secSimul}, we further compare HCT with CVT with
simulated data.

%%%%%%%%%%%%%%
%%%%%%%%%%%%%%
%%%%%%%%%%%%%%
%s1.12 #&#
\subsection{Summary and possible extensions}
\label{subsecSummary}
We propose HCT classifier for two-class classification, where the major
methodological innovation is the use of IT for feature selection and
the use of HC for threshold choice.

IT is based on an
``optimal'' linear transform that maximizes SNR in
all signal locations, and has advantages over
BT and WT.
IT also has a three-fold advantage over the well-known variable
selection methods such as the Lasso, SCAD, and Dantzig selector: (a) IT
is computationally faster, (b) IT is more approachable in terms of
delicate analysis, and (c) the tuning parameter
of IT can be conveniently set, but how to set the tuning parameters of
the other methods
remains an open problem.

The idea of using HC for threshold choice goes back to \cite
{DonohoJin2008}, where the focus is on the
case where $\Omega$ is the identity matrix (see also \cite
{HallPittelkowGhosh2008}). In this paper,
with considerable efforts, we extend the ideas to the case where
$\Omega
$ is unknown but
is presumably sparse, and show that HC achieves the optimal phase
diagram in classification.
The optimality of HC is not coincidental, and the underlying reason is
the intimate relationship between the HC functional and Fisher's
separation. This is explained in Sections~\ref{secIdealHC}--\ref
{secHCT} with details.

In Theorems \ref{thmmIdealized}--\ref{thmmUB1} and Sections~\ref{secIdealHC}--\ref{secHCT}, we assume the signals have the same signs
and strengths.
The first assumption is largely
for simplicity and can be removed.
The second assumption
can be largely relaxed, and both Theorems \ref{thmmIdealized}--\ref
{thmmUB1} and
the intimate relationship between HC and Fisher's separation continue
to hold to some extent if the signal strengths are unequal. One such
example is where
the signal distribution $H_p$, after scaled by a factor of $(\log
(p))^{-1/2}$, has a continuous density over a closed interval contained
in $(0, \infty)$ which does not depend on $p$.

In the paper, we require $\Omega$ to be $K_p$-sparse where $K_p \leq
L_p$ (see Definition~\ref{de1.2}) and does not exceed a multi-$\log(p)$ term.
This assumption is mainly used to control the
chromatic number of the induced graph $\cg(\Omega)$.
Since the chromatic number of a graph could be much smaller than
its maximum degree, the assumption on $\Omega$ can be relaxed to that of
the chromatic number of $g(\Omega)$ does not exceed a multi-$\log(p)$ term.
Also,
when $\Omega$ has many small nonzero coordinates, we can always
regularize it first with a threshold $t > 0$: $\Omega^*(i,j) = \Omega
(i,j) 1\{ |\Omega(i,j)| \geq t\}$, and the main results continue to
hold if $\Omega^*$ is $K$-sparse and the difference
between two matrices is ``sufficiently small.''

%%%%%%%%%%%%%%%
%%%%%%%%%%%%%%%
%%%%%%%%%%%%%%%
%%%%%%%%%%%%%%%
%%%%%%%%%%%%%%%
%s1.13 #&#
\subsection{Content}
The remaining part of the paper is organized as follows. In Section~\ref{secIdealHC}, we introduce two functionals: Fisher's separation and
ideal HC, and show that the two functionals are intimately connected to
each other. In Section~\ref{secHCT}, we derive a large-deviation bound
on the empirical c.d.f., and then use it to characterize the stochastic
fluctuation of the HC functional and that of Fisher's separation.
Theorems \ref{thmmIdealized}--\ref{thmmUB1} are proved at the end of
this section. All other claims (theorems and lemmas)
are proved in the supplementary material \cite{HCTghsupp}.
Section~\ref{secSimul} contains numeric examples.

%%%%%%%
%%%%%%%
%%%%%%%
%s1.14 #&#
\subsection{Notation}
In this paper, $C > 0$ and $L_p > 0$ denote a generic constant and a
generic multi-$\log(p)$ term respectively, which may vary from
occurrence to
occurrence. For two positive sequences $\{a_p\}_{p =1}^{\infty}$ and
$\{
b_p\}_{p= 1}^{\infty}$,
we say that $a_p \gtrsim b_p$ (or $a_p \lesssim b_p$) if there is a
sequence $\{\Delta_p\}_{p=1}^{\infty}$ such that $\Delta_p \goto0$ and
$a_p (1 + \Delta_p) \geq b_p$ [or $a_p (1 + \Delta_p) \leq b_p$]. We
say that
$a_p \sim b_p$ if $a_p \gtrsim b_p$ and $a_p \lesssim b_p$, and we say
that $a_p \asymp b_p$ if there is a
constant $c_0 > 1$ such that for sufficiently large $p$, $c_0^{-1}
\leq a_p / b_p \leq c_0$.

The notation $\Omega$ and $\Sigma$ are always associated with each
other by $\Omega= \Sigma^{-1}$, and $(X_i, Y_i)$ represents a training
sample while $(X, Y)$ represents a
test sample. The summarizing $z$-vector for the training data set is
denoted by $Z$, with $\tilde{Z} = \Omega Z$ and $\hat{Z} = \hat
{\Omega}
Z$, where $\hat{\Omega}$ is some estimate of $\Omega$.

%%%%%%%%%%%%%%%%%%%%%
%%%%%%%%%%%%%%%%%%%%%
%%%%%%%%%%%%%%%%%%%%%
%s2 #&#
\section{Ideal threshold and ideal HCT} \label{secIdealHC}
In Sections~\ref{secIdealHC}--\ref{secHCT}, we discuss the behavior
of HCT classifier.
We limit our discussion to
the $\operatorname{ARW}(\beta, r, \theta, \Omega)$ model, but the key ideas are
valid beyond
the $\operatorname{ARW}$ model and extensions are possible; see discussions in Section~\ref{subsecSummary}.

The key insight behind the HCT methodology is that in a broad context,
\[
\mbox{HCT $\approx$ ideal HCT $\approx$ ideal threshold}.
\]
The ideal HCT is the nonstochastic counterpart of HCT, and the
ideal threshold is the threshold one would
choose if the underlying signal structure were known.

In this section, we elaborate the intimate connection between the ideal
HCT and the ideal threshold, and their connections to Fisher's
separation. We also investigate
the performance of ``ideal classifier'' where we assume $\Omega$ is known
and the threshold is set ideally.

The connection between HCT and ideal HCT is addressed in
Section~\ref{secHCT}, which is new even in the case of $\Omega= I_p$;
compare \cite{DonohoJin2009}. Theorems \ref{thmmIdealized}--\ref
{thmmUB1} are also proved in Section~\ref{secHCT}.

%%%%%%%%%%%%%%%%%
%%%%%%%%%%%%%%%%%
%%%%%%%%%%%%%%%%%
%s2.1 #&#
\subsection{Fisher's separation and classification heuristics}
Fix a threshold $t > 0$ and let $\hat{\Omega}$ be an acceptable
estimator of $\Omega$. We are interested in the classifier that
estimates $Y = \pm1$ according to $L_t(X, \hat{\Omega})> < 0$, whereas
in (\ref{hctclassadd1})--(\ref{hctclassadd2}),
\[
L_t(X, \hat{\Omega}) = \bigl(\hat{\mu}_t^{\hat{Z}}
\bigr)' \hat{\Omega} X\qquad \mbox{where $\hat{\mu}_t^{\hat Z}(j)
= \sgn\bigl(\hat{Z}(j)\bigr) 1\bigl\{ \bigl|\hat {Z}(j)\bigr| \geq t\bigr\}$}.
\]
For any fixed $p \times1$ vector $Z$ and $p \times p$ positive
definite matrix $A$, we introduce
\[
M_p(t, Z, \mu, A) = M_p(t, Z, \mu, A; n_p) =
\bigl(\hat{\mu}_t^Z\bigr)' A \mu
\]
and
\[
V_p(t, Z, A) = V_p(t, Z, A; \Omega) = \bigl(\hat{
\mu}_t^Z\bigr)' A \Omega ^{-1} A
\hat{\mu}_t^Z,
\]
where loosely, ``$M$'' and ``$V$'' stand for the mean and variance, respectively.
In our model, given $(\mu, \hat{Z}, \hat{\Omega})$, the test sample $X
\sim N(Y \cdot\mu, \Omega^{-1})$; see (\ref{testsample}) and note that
$\hat{\Omega}$ is independent of $X$ since it is acceptable. It
follows that
\[
L_t(X, \hat{\Omega})\sim N \bigl( Y \cdot M_p(t,
\hat{Z}, \mu, \hat {\Omega}), V_p(t, \hat{Z}, \hat{\Omega}) \bigr),
\]
and the misclassification error rate of $L_t(X, \hat{\Omega})$ is
%
%e2.1 #&#
%
\begin{equation}
\label{misclassadd1} P\bigl( Y \cdot L_t(X, \hat{\Omega}) < 0 |
\mu, \hat{Z}, \hat{\Omega}\bigr) = \bphi \biggl( \frac{M_p(t, \hat Z, \mu, \hat\Omega)}{\sqrt{V_p(t,
\hat
Z, \hat{\Omega})}} \biggr),
\end{equation}
where $\bphi= 1 - \Phi$ denotes the survival function of $N(0,1)$.

The right-hand side of (\ref{misclassadd1}) is closely related to the
well-known Fisher's separation (Sep) \cite{Anderson2003}, which
measures the standardized interclass distance
$\operatorname{Sep}(t, \hat{Z}, \mu,  \hat{\Omega}) = \operatorname{Sep}(t, \hat{Z}, \mu, \hat
{\Omega
}; \Omega, p)$:
%
%e2.2 #&#
%
\begin{equation}
\label{DefineSep}\qquad \operatorname{Sep}(t, \hat{Z}, \mu, \hat{\Omega}; \Omega, p) =
\frac{E[L_t(X,
\hat
{\Omega}) | Y = 1] - E[L_t(X, \hat{\Omega}) | Y = -1] }{SD(L_t(X,
\hat
{\Omega}))}.
\end{equation}
In fact, it is seen that
$\operatorname{Sep}(t, \hat{Z}, \mu, \hat{\Omega}) = 2M_p(t, \hat{Z}, \mu, \hat
{\Omega
}) / \sqrt{V_p(t, \hat{Z}, \hat{\Omega})}$,
and (\ref{misclassadd1}) can be rewritten as
\[
P\bigl( Y \cdot L_t(X, \hat\Omega) < 0 | \mu, \hat Z, \hat{\Omega}
\bigr) = \bphi \bigl(\tfrac{1}{2} \operatorname{Sep}(t, \hat Z, \mu, \hat{\Omega}) \bigr).
\]
By (\ref{DefineH}) and (\ref{pointmass}), the overall misclassification
error rate is then
%
%e2.3 #&#
%
\begin{equation}
\label{Definemisclassification} P\bigl(Y \cdot L_t(X, \hat{\Omega})
< 0\bigr) = E_{\eps_p, \tau_p} E \bigl[ \bphi \bigl(\tfrac{1}{2} \operatorname{Sep}(t,
\hat{Z}, \mu, \hat{\Omega}) \bigr) \bigr],
\end{equation}
where $E$ is the expectation with respect to the law of $(\hat{Z},
\hat
{\Omega} | \mu)$, and
$E_{\eps_p, \tau_p}$ is the expectation with respect to the law of
$\mu
$; see (\ref{DefineH}) and (\ref{pointmass}).

We introduce two proxies for Fisher's separation. Throughout this paper,
%
%e2.4 #&#
%
\begin{equation}
\label{DefinetildeZ} \tilde{Z} = \Omega Z.
\end{equation}
For the first proxy, recall that $\hat{Z} = \hat{\Omega} Z$ [e.g.,
(\ref
{DefinehatZ})]. Heuristically, $\hat{\Omega} \approx\Omega$ and so
$\hat{Z} \approx\tilde{Z}$.
We expect that $\operatorname{Sep}(t, \hat Z, \mu, \hat{\Omega}) \approx \operatorname{Sep}(t,
\tilde
{Z}, \mu, \Omega)$; the latter is Fisher's separation
for the idealized case where $\Omega$ is known and is defined as
%
%e2.5 #&#
%
\begin{equation}
\label{sepproxy} \operatorname{Sep}(t, \tilde{Z}, \mu, \Omega) = 2 M_p(t,
\tilde{Z}, \mu, \Omega) / \sqrt{V_p(t, \tilde{Z}, \Omega)}.
\end{equation}
For the second proxy, we note that when $p$ is large, some regularity
appears, and we expect that $M_p(t, \tilde{Z}, \mu, \Omega) \approx
m_p(t, \eps_p, \tau_p, \Omega)$ and $V_p(t, \tilde{Z}, \Omega)
\approx
v_p(t, \eps_p, \tau_p, \Omega)$, where
%
%e2.6 #&#
%
\begin{eqnarray}
\label{Definesmallmv} m_p(t, \eps_p, \tau_p,
\Omega) &=& E\bigl[M_p(t, \tilde{Z}, \mu, \Omega)\bigr],
\nonumber
\\[-8pt]
\\[-8pt]
\nonumber
v_p(t, \eps_p, \tau_p, \Omega) &= &E
\bigl[V_p(t, \tilde{Z}, \Omega)\bigr].
\end{eqnarray}
In light of this, a second proxy separation is the \textit{population $\operatorname{Sep}$}:
\[
\widetilde{\operatorname{Sep}}(t) = \widetilde{\operatorname{Sep}}(t, \eps_p, \tau_p,
\Omega) = 2 m_p(t, \eps_p, \tau_p, \Omega) /
\sqrt{v_p(t, \eps_p, \tau_p, \Omega)}.
\]
In summary, we expect to see that
\[
\operatorname{Sep}(t, \hat{Z}, \mu, \hat{\Omega}) \approx \operatorname{Sep}(t, \tilde{Z}, \mu, \Omega)
\approx\widetilde{\operatorname{Sep}}(t, \eps_p, \tau_p, \Omega),
\]
and that
%
%e2.7 #&#
%
\begin{equation}
\label{errorapprox} P\bigl(Y \cdot L_t(X, \hat{\Omega}) < 0\bigr)
\approx\bphi \bigl(\tfrac{1}{2} \widetilde{\operatorname{Sep}}(t) \bigr).
\end{equation}
In Section~\ref{secHCT}, we solidify the above connections. But before
we do that, we study the ideal threshold---the threshold that maximizes
$\widetilde{\operatorname{Sep}}(t)$.
%%%%%%%%%%%%%%%
%%%%%%%%%%%%%%%
%%%%%%%%%%%%%%%
%%%%%%%%%%%%%%%
%%%%%%%%%%%%%%%
%s2.2 #&#
\subsection{Ideal threshold}
\label{subsecidealtadd}
Ideally, one would choose $t$ to minimize the classification error of
$L_t(X, \hat{\Omega})$. In light of (\ref{errorapprox}), this is almost
equivalent to choosing $t$ as the ideal threshold.
%
%de2.1 #&#
\begin{definition}\label{de2.1}
The ideal threshold $T_{\mathrm{ideal}}(\eps_p, \tau_p, \Omega)$ is the
maximizing point of the second proxy: $T_{\mathrm{ideal}}(\eps_p, \tau_p,
\Omega
) = \margmax_{\{0 < t < \infty\}} \widetilde{\operatorname{Sep}}(t, \eps_p, \tau_p,
\Omega)$.
\end{definition}
In general, $\widetilde{\operatorname{Sep}}(t, \eps_p, \tau_p, \Omega)$ and
$T_{\mathrm{ideal}}(\eps_p, \tau_p, \Omega)$ may depend on $\Omega$ in a
complicated way. Fortunately, it turns out that for large $p$ and all
$\Omega$ in ${\cal M}_p^*(a, K_p)$ [see (\ref{eqdefMset0})], the
leading terms of $\widetilde{\operatorname{Sep}}(t)$ and $T_{\mathrm{ideal}}(\eps_p, \tau_p,
\Omega)$ do not depend on the off-diagonals of $\Omega$ and have rather
simple forms.
%
%de2.2 #&#
\begin{definition}[(Folding)]\label{de2.2}
Denote $\Psi_{\tau}(t) = P( |N(\tau, 1)| \leq t)$.
When $\tau= 0$, we drop the subscript and write $\Psi(t)$. Also,
denote $\bar{\Psi}_{\tau} = 1 - \Psi_{\tau}(t)$ and $\bar{\Psi
}(t) = 1
- \Psi(t)$.
\end{definition}
In detail, let
%
%e2.8 #&#
%
\begin{eqnarray}
\label{tw0} \widetilde{W}_0(t)& =& \widetilde{W}_0(t,
\eps_p, \tau_p; \Psi) = \eps_p
\bpsi_{\tau_p} (t) / \sqrt{\bpsi(t) + \eps_p
\bpsi_{\tau_p}(t)},\\
%%
%%%%%%%%
%%%%%%%%
%%%%%%%%
%%
%%e2.9 #&#
\label{eqdeftp^} t_p^*(\beta, r)& =& \min \biggl\{2,
\frac{r + \beta}{2r} \biggr\} \tau_p
\end{eqnarray}
and
%%%%%%%%%%%%
%%%%%%%%%%%%
%%%%%%%%%%%%
%
%e2.10 #&#
%
\begin{equation}
\label{eqdefdelta} \delta(\beta, r) = \cases{ %
\beta- r, &\quad  $r \leq\beta/3$,
\vspace*{2pt}\cr
\displaystyle\frac{(\beta+r)^2}{8r}, & \quad$\beta/3 < r < \beta$,
\vspace*{2pt}\cr
\beta/2, & \quad$\beta\leq r < 1.$}
\end{equation}
Elementary calculus shows that for large $p$,
%%%%%%%%%%%
%%%%%%%%%%%
%%%%%%%%%%%
%%%%%%%%%%%
%
%e2.11 #&#
%
\begin{equation}
\label{eqoptimizeW0} \mathop{\margmax}_{\{ 0 \leq t < \infty\}} \bigl\{ \widetilde{W}_0(t)
\bigr\} \sim t_p^*(\beta, r),\qquad \sup_{\{ 0 \leq t < \infty\}} \widetilde
{W}_0(t) = L_p \cdot p^{-\delta(\beta, r)}.
\end{equation}
%
%%%%%%%%
%%%%%%%%
%%%%%%%%
%%%%%%%%
It turns out that there is an intimate relationship between $\widetilde
{\operatorname{Sep}}(t, \eps_p, \tau_p, \Omega)$ and
$\widetilde{W}_0(t, \eps_p, \tau_p)$, where the latter does not depend
on the off-diagonals of $\Omega$. To see the point, we discuss the
cases $r < \beta$ and $r \geq\beta$ separately.

In the first case, for $a$ as in ${\cal M}_p^*(a, K_p)$, we let
%
%e2.12 #&#
%
\begin{eqnarray}
\label{eqdefc0} c_0(\beta,r,a) &=& \delta\bigl(\beta,
a^2r\bigr)- \delta(\beta, r),
\nonumber
\\[-8pt]
\\[-8pt]
\nonumber
\tilde c_0(\beta,r,a)& =&
\tilde c_1(\beta,r,a) - \delta(\beta, r),
\end{eqnarray}
where $c_0(\beta,r,a)>0$ for $r<\beta$; if $a < 1/3$, $\tilde
c_1(\beta
,r,a) = \beta$, and otherwise,
\[
\tilde c_1(\beta,r,a) = \cases{
\displaystyle\frac{(3a-1) r}{3 - a}+\beta, &\quad $\displaystyle r\leq\frac{3-a}{1+5a}\beta$,
\vspace*{2pt}\cr
\displaystyle\frac{3-a}{1+a}\frac{(\beta+r)^2}{8r}, &\quad $\displaystyle \frac{3-a}{1+5a}\beta< r <
\beta.$}
\]
The following lemma is proved in the supplementary material \cite{HCTghsupp}.
%%%%%%%%%%%%%%
%%%%%%%%%%%%%%
%%%%%%%%%%%%%%
%%%%%%%%%%%%%%
%
%le2.1 #&#
\begin{lemma}\label{lemmaidealsep}
Fix $(\beta, r, \theta, a) \in(0,1)^4$ such that $\rho_\theta
^*(\beta
)<r < \beta$ and $(1-\theta)/2<\beta< (1-\theta)$. In the
$\operatorname{ARW}(\beta,
r, \theta, \Omega)$ model, as $p \goto\infty$,
\begin{eqnarray*}
&& \sup_{t >0} \sup_{\{\Omega\in{\cal M}_p^*(a,K_p) \}}\bigl | p^{
{(\theta-1)}/{2}}
\widetilde{\operatorname{Sep}}(t, \eps_p, \tau_p, \Omega) - 2
\tau_p\tw _0(t, \eps_p, \tau_p) \bigr|\\
&&\qquad\leq L_p p^{-\max\{\beta-
{r}/{2}, {(3\beta+r)}/{4}\}}\\
&&\qquad\quad{}+ L_p \bigl[p^{-\min\{r,{(\beta-r)}/{2},(1-a)(\beta-ar)\}}+p^{-
c_0(\beta,r,a)} + p^{-\tilde c_0(\beta,r,a)}
\bigr]\\
&&\qquad\quad{}\times \sup_{\{0< t <
\infty
\}} \tw_0(t, \eps_p,
\tau_p).
\end{eqnarray*}
\end{lemma}
Note that $\delta(\beta,r) < \max(\beta- r/2, (3\beta+ r)/4)$. As a
result, approximately, $\widetilde{\operatorname{Sep}}(t, \eps_p, \tau_p,\Omega)
\propto\widetilde{W}_0(t, \eps_p, \tau_p)$ for all $\Omega\in
{\cal
M}_p^*(a, K_p)$. Combining this\break with~(\ref{eqoptimizeW0}), we expect
to have
%
%e2.13 #&#
%
\begin{eqnarray}
\label{eqidealapprox} T_{\mathrm{ideal}}(\eps_p, \tau_p,
\Omega) &\sim& t_p^*(\beta, r),
\nonumber
\\[-8pt]
\\[-8pt]
\nonumber
\sup_{0 < t
< \infty}
\widetilde{\operatorname{Sep}}(t, \eps_p, \tau_p, \Omega) &=&
L_p p^{
{(1-\theta)}/{2}-\delta(\beta, r)}.
\end{eqnarray}

Next, consider the case $r \geq\beta$. The lemma below is proved in
the supplementary material \cite{HCTghsupp}.
%%%%%%%%%%%%%%%%%
%%%%%%%%%%%%%%%%%
%%%%%%%%%%%%%%%%%
%%%%%%%%%%%%%%%%%
%
%le2.2 #&#
\begin{lemma}\label{lemmaidealsepb}
Fix $(\beta, r, \theta, a) \in(0,1)^4$ such that $r \geq\beta$ and
$(1-\theta)/2<\beta<(1-\theta)$. Let $\Delta_1 = d_0\log(\log
(p))/\sqrt {\log p}$ and $\Delta_2 = 2\sqrt{\log(K_p \log p)}$, where $d_0 > 0$ is
some constant. In the $\operatorname{ARW}(\beta, r, \theta, \Omega)$ model with
$\Omega\in\widetilde{\cal M}_p^*(a,b,K_p)$, as $p \goto\infty$,
\begin{longlist}[(a)]
\item[(a)] $\sup_{\{0 < t < \sqrt{2 \beta\log(p)} - \Delta_1 \}}
\widetilde{\operatorname{Sep}}(t, \eps_p, \tau_p, \Omega)\lesssim\frac{5}{3}\tau
_pK_p^{-1} p^{(1-\theta-\beta)/2}$,\eject
\item[(b)] $\sup_{\{t \geq\tau_p + \Delta_2 \}} \widetilde
{\operatorname{Sep}}(t, \eps
_p, \tau_p, \Omega)\lesssim\frac{5}{3}\tau_pK_p^{-1} p^{(1- \theta
-\beta)/2}$,
\item[(c)] $\sup_{\{ \sqrt{2 \beta\log p} - \Delta_1 \leq t < \tau
_p \}
} \widetilde{\operatorname{Sep}}(t, \eps_p, \tau_p, \Omega) \gtrsim2 \tau_p K_p^{-1}
p^{{(1-\theta-\beta)}/{2}}$ and \\ $ \sup_{\{ t>0 \}} \widetilde
{\operatorname{Sep}}(t, \eps_p, \tau_p, \Omega) \leq L_pp^{{(1-\theta-\beta)}/{2}}$.
\end{longlist}
\end{lemma}
A direct result of Lemma~\ref{lemmaidealsepb} is that, for all
$\Omega
\in\widetilde{\cal M}_p^*(a,b,K_p)$ [see (\ref{eqdefMset0})],
%
%e2.14 #&#
%
\begin{eqnarray}
\label{eqidealsep} \sqrt{2 \beta\log(p)} &\lesssim& T_{\mathrm{ideal}} \lesssim
\sqrt{2 r \log(p)},
\nonumber
\\[-8pt]
\\[-8pt]
\nonumber
\sup_{\{ 0 < t < \infty\} } \bigl\{ \widetilde{\operatorname{Sep}}(t) \bigr
\} &\asymp& L_pp^{(1-\theta-\beta)/2},
\end{eqnarray}
where $T_{\mathrm{ideal}} = T_{\mathrm{ideal}}(\eps_p, \tau_p, \Omega)$ and
$\widetilde
{\operatorname{Sep}}(t) = \widetilde{\operatorname{Sep}}(t, \eps_p, \tau_p, \Omega)$ for short. In
this case, the function $\widetilde{\operatorname{Sep}}(t)$
sharply increases and decreases in the intervals $(0, \sqrt{2 \beta
\log
(p)})$ and $(\sqrt{2 r \log(p)}, \infty)$, respectively, but is
relatively flat in the interval $(\sqrt{2 \beta\log(p)}, \sqrt{2 r
\log
(p)})$; in this interval, the function reaches the maximum but
varies slowly at the magnitude of $O(L_pp^{(1 -\theta-\beta)/2})$. In
the current case, on one hand, it is not critical to pin down
$T_{\mathrm{ideal}}$, as $\widetilde{\operatorname{Sep}}(t) = L_pp^{(1-\theta-\beta)/2}$ for
all~$t$ in the whole interval.
On the other hand, it is hard to pin down $T_{\mathrm{ideal}}$ uniformly for all
$\Omega$ under consideration, if possible at all.
%%%%%%%%%%%%%%
%%%%%%%%%%%%%%
%%%%%%%%%%%%%%
%%%%%%%%%%%%%%
%s2.3 #&#
\subsection{Ideal HCT} Ideal HCT is a counterpart of HCT and a
nonstochastic threshold that HCT tries to estimate.
Introduce a functional which is defined over all survival functions
associated with a positive random variable:
\[
\operatorname{HC}(t, G) = \sqrt{p}\bigl[G(t) - \bar{\Psi}(t)\bigr] / \sqrt{G(t) \bigl(1 - G(t)
\bigr)},\qquad t > 0.
\]
We are primarily interested in thresholds that are neither too small or
too large as far as HCT concerns; see (\ref{Definesnp}).
In light of this, we introduce the HCT functional
\[
T_{\mathrm{HC}}(G) = \mathop{\margmax}_{\{\bpsi^{-1}({1}/{2})<t < s_p^*\}} \operatorname{HC}(t, G),
\]
where the term $\bpsi^{-1}(1/2)$ is chosen for convenience, and can be
replaced by some other positive constants.
Recall that $\tilde{Z} = \Omega Z$ and $\hat{Z} = \hat{\Omega} Z$
[e.g., (\ref{DefinetildeZ}) and (\ref{DefinehatZ})]. For any $t > 0$, let
%
%e2.15 #&#
%
\begin{equation}
\label{Definecf} \cf_p(t) = \frac{1}{p} \sum
_{j = 1}^p 1\bigl\{ \bigl|\hat{Z}(j)\bigr| \geq t \bigr\}
\end{equation}
and
%
%e2.16 #&#
%
\begin{eqnarray}
\label{Definetf} \tf_p(t) &=& \frac{1}{p} \sum
_{j = 1}^p 1\bigl\{ \bigl|\tilde{Z}(j)\bigr| \geq t \bigr\},
\nonumber
\\[-8pt]
\\[-8pt]
\nonumber
\tf(t) &=& \tf(t, \eps_p, \pi_p, \Omega) = E_{\eps_p, \pi_p}
\bigl[ \tf_p(t)\bigr].
\end{eqnarray}
Note that the only difference between $\tf_p(t)$ and $\tf(t)$ is the
subscript $p$.
Heuristically, for large $p$, we expect to have
$\cf_p(t) \approx\tf_p(t) \approx\tf(t)$.
As a result, we expect that
\[
T_{\mathrm{HC}}(\cf_p) \approx T_{\mathrm{HC}}(\tf_p)
\approx T_{\mathrm{HC}}(\tf),
\]
where $T_{\mathrm{HC}}(\cf_p)$ is the
HCT where $\Omega$ is unknown and has to be estimated, $T_{\mathrm{HC}}(\tf_p)$
is the HCT when $\Omega$ is known, and $T_{\mathrm{HC}}(\tf)$ is a
nonstochastic counterpart of $T_{\mathrm{HC}}(\tf_p)$.
Note that in disguise, $T_{\mathrm{HC}}(\cf_p)$ is the same as $t_{\mathrm{HC}}^*$, the
HCT defined in (\ref{HCTrefined}).
%
%de2.3 #&#
\begin{definition}\label{de2.3}
We call $T_{\mathrm{HC}}(\tf)$ the ideal Higher Criticism Threshold (ideal HCT).
\end{definition}
Similarly, the leading term of ideal HCT has a simple form that is easy
to analyze. Fix $1 \leq j \leq p$. Let $D_j = \{k\dvtx 1 \leq k \leq p,
\Omega(j, k) \neq0\}$, and let
\begin{eqnarray*}
g_1(t) &=& g_1(t; \Omega, \eps_p,
\tau_p)\\
 &=& \frac{1}{p} \sum_{j = 1}^p
P \bigl(\bigl|\tilde{Z}(j)\bigr| \geq t, \mu(k) \neq0 \mbox{ for some $k \in
D_j$, $k\neq j$} \bigr).
\end{eqnarray*}
The following is a counterpart of $\tw_0(t)$ defined in (\ref{tw0}) and
can be well approximated by the latter:
%
%e2.17 #&#
%
\begin{equation}
\label{eqdefineW0t} W_0(t) = W_0(t,
\eps_p, \tau_p, \Omega) = \frac{\eps_p \bpsi
_{\tau
_p}(t) + g_1(t)}{\sqrt{\bpsi(t) + \eps_p \bpsi_{\tau_p}(t) + g_1(t)}}.
\end{equation}
The following lemmas are proved in the supplementary material \cite{HCTghsupp}.
%%%%%%%%%%%%%%%
%%%%%%%%%%%%%%%
%%%%%%%%%%%%%%%
%
%le2.3 #&#
\begin{lemma} \label{lemmaIdealHCT}
Fix $(\beta, r, \theta, a) \in(0, 1)^4$ such that $r > \rho_{\theta
}^*(\beta)$ and $(1-\theta)/2 < \beta< (1-\theta)$. In the
$\operatorname{ARW}(\beta
,r,\theta, \Omega)$ model, as $p \goto\infty$,
\[
\sup_{\{t > \bpsi^{-1}({1}/{2}) \}} \sup_{\{ \Omega\in{\cal
M}^*_p(a, K_p) \}} \bigl\{
\bigl|p^{-1/2}\operatorname{HC}(t,\tf)-W_0(t,\eps_p,\tau
_p,\Omega ) \bigr| \bigr\} \leq L_p p^{- \beta}.
\]
\end{lemma}
%
%%%%%%%%%%%%%
%%%%%%%%%%%%%
%%%%%%%%%%%%%
%%%%%%%%%%%%%
%
%le2.4 #&#
\begin{lemma} \label{lemmaIdealHCT2}
Fix $(\beta, r, \theta, a) \in(0, 1)^4$ such that $r > \rho_{\theta
}^*(\beta)$ and $(1-\theta)/2 < \beta< (1-\theta)$. In the
$\operatorname{ARW}(\beta
,r,\theta, \Omega)$ model, as $p \goto\infty$, we have
\begin{eqnarray*}
&&\sup_{\{t > 0 \}} \sup_{\{ \Omega\in{\cal M}_p^*(a,K_p) \}} \bigl|W_0(t,
\eps_p, \tau_p, \Omega) - \tw_0(t,
\eps_p, \tau_p)\bigr |\\
&&\qquad \leq L_p
\Bigl[p^{-{3\beta}/{2}} + p^{- c_0(\beta, r, a)} \sup_{\{t
> 0
\}}
\tw_0(t) \Bigr].
\end{eqnarray*}
If additionally $r \geq\beta$, then:
\begin{longlist}[(a)]
\item[(a)] $\sup_{\{0 \leq t < \sqrt{2 \beta\log(p)} - \Delta_1 \}}
W_0(t, \eps_p, \tau_p, \Omega) \lesssim(\frac{1}{\sqrt{2}})
p^{-\beta/2}$,\eject
\item[(b)] $\sup_{\{\tau_p \leq t < \infty\}} W_0(t, \eps_p, \tau_p,
\Omega) \lesssim(\frac{1}{\sqrt{2}}) p^{-\beta/2}$,
\item[(c)] $\frac{3}{4} p^{-\beta/2}\lesssim\sup_{\{\sqrt{2 \beta
\log
(p)} - \Delta_1 < t < \tau_p \}} W_0(t, \eps_p, \tau_p, \Omega)
\leq
L_pp^{-\beta/2} $,
\end{longlist}
where $\Delta_1 = d_0 \log\log(p)/ \sqrt{\log(p)}$ is defined in Lemma~\ref{lemmaidealsepb}.
\end{lemma}
Lemmas \ref{lemmaIdealHCT}--\ref{lemmaIdealHCT2} say that,
approximately, $\operatorname{HC}(t, \tf) \propto W_0(t)$, and that two functions
$\tw
_0(t)$ and $W_0(t)$ are generally close.

%s2.4 #&#
\subsection{Relationship between two ideal thresholds and
classification by the ideal classifier}
Together, Lemmas \ref{lemmaidealsep}--\ref{lemmaIdealHCT2}
consolidate the intimate relationship
between the ideal threshold and the ideal HCT. To see the point, we
discuss the cases $r < \beta$ and $r \geq\beta$ separately.

For the first case, write $T_{\mathrm{ideal}} = T_{\mathrm{ideal}}(\eps_p, \tau_p,
\Omega
)$ and
$\widetilde{\operatorname{Sep}}(t) = \widetilde{\operatorname{Sep}}(t, \eps_p, \tau_p, \Omega)$ for
short as before. The following theorem is proved in the supplementary
material~\cite{HCTghsupp}.
%%%%%%%%%%%%%
%%%%%%%%%%%%%
%%%%%%%%%%%%%
%%%%%%%%%%%%%
%
%th2.1 #&#
\begin{thmm} \label{thmmIdealHCTa}
Fix $(\beta, r, \theta, a) \in(0,1)^4$ such that $\rho_{\theta
}^*(\beta
) < r < \beta$ and $(1-\theta)/2 < \beta< (1-\theta)$. In the
$\operatorname{ARW}(\beta,r,\theta, \Omega)$ model with\vspace*{1pt} $\Omega\in\mathcal
{M}_p^*(a,K_p)$, as $p \goto\infty$, there is a constant $c_1 =
c_1(\beta, r, a) > 0$ such that
$|T_{\mathrm{HC}}(\tf) - T_{\mathrm{ideal}}| \leq L_p p^{ - c_1(\beta, r, a)}$, and so
$\widetilde{\operatorname{Sep}}(T_{\mathrm{HC}}(\tf)) \sim\widetilde{\operatorname{Sep}}(T_{\mathrm{ideal}}) = L_p
p^{(1 - \theta)/2 - \delta(\beta, r)}$.
\end{thmm}

Consider the second case. Lemma~\ref{lemmaIdealHCT2} says that
$\sqrt{2 \beta\log(p)} \lesssim T_{\mathrm{HC}}(\tf) \lesssim\sqrt{2 r
\log(p)}$.
While it is hard to further elaborate how close two ideal thresholds
are, in light of (\ref{eqidealsep}), HC classification with any $t$ in
this range is successful,
so it is not critical to pin down the ideal HCT. The following theorem
is proved in
the supplementary material \cite{HCTghsupp}.
%%%%%%%%%%
%%%%%%%%%%
%%%%%%%%%%
%%%%%%%%%%
%
%th2.2 #&#
\begin{thmm} \label{thmmIdealHCTb}
Fix $(\beta, r, \theta, a) \in(0,1)^4$ such that $r \geq\beta$ and
$(1-\theta)/2 < \beta< (1-\theta)$. In the $\operatorname{ARW}(\beta, r, \theta, a)$
model where $\Omega\in\widetilde{\mathcal{M}}_p^*(a,b,K_p)$, as $p
\goto\infty$, we have that
$2\tau_pK_p^{-1}p^{(1-\theta-\beta)/2}\lesssim\widetilde
{\operatorname{Sep}}(T_{\mathrm{HC}}(\tf)) \leq\widetilde{\operatorname{Sep}}(T_{\mathrm{ideal}}(\eps_p,\tau_p,
\Omega
)) = L_pp^{(1-\theta-\beta)/2}$.
\end{thmm}

To conclude this section, we investigate the ``ideal'' classifier $L_t(X,
\Omega)$, where $\Omega$ is known to us.
Note that for each fixed $t$, the misclassification error of $L_t(X,
\Omega)$ is
$P( Y\cdot L_t(X,\Omega)< 0) = E_{\eps_p, \pi_p} E  [ \bphi(
\frac
{1}{2}\operatorname{Sep}(t, \tilde{Z}, \mu, \Omega)  ]$.
The following theorem is proved in the supplementary material \cite{HCTghsupp}.
%%%%%%%%%%%%%%%%%%
%%%%%%%%%%%%%%%%%%
%%%%%%%%%%%%%%%%%%
%%%%%%%%%%%%%%%%%%
%
%th2.3 #&#
\begin{thmm} \label{thmmsep}
Fix $(\beta, r, \theta, a) \in(0,1)^4$ such that $(1 - \theta)/2 <
\beta< (1 - \theta)$ and $r > \rho_{\theta}^*(\beta)$. In the
$\operatorname{ARW}(\beta, r, \theta, a)$ model with $\Omega\in\widetilde{\cal
M}_p^*(a,b,K_p)$, as $p \goto\infty$,
\[
\min_{t} P \bigl(Y\cdot L_t(X,\Omega)<0 | t
\bigr) = \bphi \bigl( \bigl(1 + o(1)\bigr) \cdot\tfrac{1}{2}\widetilde{\operatorname{Sep}}
(T_{\mathrm{ideal}} ) \bigr).
\]
When $r <\beta$, the condition $\Omega\in\widetilde{\cal
M}_p^*(a,b,K_p)$ can be relaxed to that of $\Omega\in{\cal M}_p^*(a, K_p)$.
\end{thmm}

Combining Theorem~\ref{thmmsep} with Theorems \ref
{thmmIdealHCTa}--\ref
{thmmIdealHCTb},
\[
\min_{t} P \bigl(Y\cdot L_t(X,\Omega)<0 | t
\bigr) = \bphi \bigl( h(t) \cdot\widetilde{\operatorname{Sep}} \bigl(T_{\mathrm{HC}}(\tf) \bigr)
\bigr),
\]
where $h(t) = h(t; \beta, r, \theta, a, \Omega_p, p)$ satisfies
$h(t) =
1/2 + o(1)$ when $r < \beta$ and $h(t) = L_p$ when $r \geq\beta$.
Recall that in both cases, $\widetilde{\operatorname{Sep}}(T_{\mathrm{ideal}}) = L_p
\widetilde
{\operatorname{Sep}}(T_{\mathrm{HC}}(\tf)) = L_p p^{(1 - \theta)/2 - \delta(\beta, r)}$, where
the exponent $(1 - \theta)/2 - \delta(\beta, r)$ is strictly positive
by the assumption of $r > \rho_{\theta}^*(\beta)$.
Therefore, if $(\beta, r)$ fall in Region of Possibility and if we set
$t$ as either of the two ideal thresholds, then $L_t(X, \Omega)$ not
only gives successful classification, but the classification error
converges to $0$ very fast.

%%%%%%%%%%%%%%%%%%
%%%%%%%%%%%%%%%%%%
%%%%%%%%%%%%%%%%%%
%%%%%%%%%%%%%%%%%%
%s3 #&#
\section{Classification by HCT}
\label{secHCT}
In the preceding section, we have been focusing on two ideal
thresholds. In this section, we study the empirical quantities, and
characterize the stochastic fluctuation of HCT and Sep
defined in (\ref{DefineSep}). We conclude the section by proving Theorems
\ref{thmmIdealized}--\ref{thmmUB1}. The main results in this section
are new, even in the idealized case where $\Omega= I_p$.
%%%%%%%%%%%%%%%
%%%%%%%%%%%%%%%
%%%%%%%%%%%%%%%
%%%%%%%%%%%%%%%
%s3.1 #&#
\subsection{Stochastic control on the HC functional} \label{subsecHCT}
Recall that
\[
\operatorname{HC}(t, \cf_p) = \sqrt{p} \bigl[\cf_p(t) - \bar{\Psi}(t)
\bigr] / \sqrt{\cf_p(t) \bigl(1 - \cf_p(t)\bigr)}.
\]
When $\cf_p(t) = 0$, the above is not well defined, and we modify the
definition slightly by replacing $\cf_p(t)$ with $1/p$. The change does
not affect the proof of the results.
The stochastic fluctuation of HCT comes from that of $\cf_p(t)$, which
consists of two components: that of estimating $\Omega$ and that of the
data. This is captured in the following triangle inequality [see (\ref
{Definecf})--(\ref{Definetf})]:
\[
\bigl| \cf_p(t) - \tf(t)\bigr| \leq\bigl|\tf_p(t) - \tf(t)\bigr| +\bigl |
\cf_p(t) - \tf_p(t)\bigr|.
\]

Consider $|\tf_p(t) - \tf(t)|$ first. The key is to study
\[
\sqrt{p} \bigl( \tf_p(t) - \tf(t) \bigr)/ \sqrt{\tf(t) \bigl(1 -
\tf(t)\bigr)}.
\]
When $\Omega= I_p$, this is the \textit{standard uniform stochastic
processes} \cite{ShorackWellner1986}
and much is known about its stochastic fluctuation. In the more general
case where
$\Omega\neq I_p$, it is usually hard to derive a tight bound on the
tail probability of this process.
Fortunately, when $\Omega$ is $K_p$-sparse, tight bounds are possible,
and the key is graph theory on the chromatic number introduced in Lemma~\ref{lemmapartition}.

Recall that $s_p^* = \sqrt{2 \log(p)}$ [e.g., (\ref{Definesnp})]. The
following lemma is the direct result of Lemma~\ref{lemmapartition}
and the
well-known Bennet's inequality \cite{ShorackWellner1986}, and is
proved in
the supplementary material \cite{HCTghsupp}.
%%%%%%%%%%%%%%%
%%%%%%%%%%%%%%%
%%%%%%%%%%%%%%%
%%%%%%%%%%%%%%%
%
%le3.1 #&#
\begin{lemma} \label{lemmastochunif}
Fix $(\beta, r, \theta, a) \in(0,1)^4$ and consider an $\operatorname{ARW}(\beta, r,
\theta, \Omega)$ model with $\Omega\in{\cal M}_p^*(a, K_p)$. As $p
\goto\infty$, there is a constant $C > 0$ such that with probability
at least $1 - o(p^{-1})$, for all $t$ satisfying $\bpsi^{-1}(1/2) < t
< s_p^*$,
\[
\sqrt{p} \bigl|\tf_p(t) - \tf(t)\bigr| / \sqrt{\tf(t) \bigl(1 - \tf(t)\bigr)}
\leq C K_p^3 \bigl(\log(p) \bigr)^{7/4}.
\]
\end{lemma}

Next, consider $|\tf_p(t) - \cf_p(t)|$. Recall that $n_p = p^{\theta}$.
By definition, if $\hat{\Omega}$ is an acceptable estimator of
$\Omega
$, then there is a constant $C > 0$ such that with probability at least
$1 - o(p^{-1})$,
%
%e3.1 #&#
%
\begin{equation}
\label{hatomega} \max_{\{ 1 \leq i,j \leq p\}} \bigl\{ \bigl| \hat{\Omega}(i,j) -
\Omega (i,j)\bigr | \bigr\} \leq CK_p^2 \sqrt{2 \log(p)} \cdot
p^{-\theta/2}.
\end{equation}
As a result, we have the following lemma, whose proof is
straightforward and thus omitted. Recall that $\hat{Z} = \hat{\Omega}
Z$ and $\tilde{Z} = \Omega Z$ [e.g., (\ref{DefinehatZ}) and (\ref
{DefinetildeZ})].
%%%%%%%%%%%%%
%%%%%%%%%%%%%
%%%%%%%%%%%%%
%
%le3.2 #&#
\begin{lemma} \label{lemmahatomega}
For any acceptable estimator $\hat{\Omega}$, $\max_{\{1 \leq j \leq
p \}
}  \{ |\hat{Z}(j) -\break \tilde{Z}(j)|  \} \leq C K_p^3 \log(p)
p^{-\theta/2}$ with probability at least $1 - o(1/p)$.
\end{lemma}
Write for short $\eta_p = C K_p^3 \log(p) p^{-\theta/2}$.
By Lemma~\ref{lemmahatomega}, with probability at least $1 - o(1/p)$,
for all $1 \leq j \leq p$,
$ | 1\{ |\hat{Z}(j)| \geq t \} - 1\{ |\tilde{Z}(j)| \geq t\}
|
\leq1\{{ t - \eta_p \leq|\tZ(j)| \leq t + \eta_p\}}$.
As a result,
\[
\bigl|\tf_p(t) - \cf_p(t)\bigr| \leq\tf_p(t -
\eta_p) - \tf_p(t + \eta_p),
\]
where we note that heuristically,
\[
\tf_p(t - \eta_p) - \tf_p(t +
\eta_p) \approx \tf(t - \eta_p) - \tf(t +
\eta_p) \approx2 \eta_p \bigl|\tf'(t)\bigr|.
\]
Combining these, with probability at least $1 - o(1/p)$, for any
$t>\bpsi^{-1}(\frac{1}{2})$,
\[
\frac{\sqrt{p}|\tf_p(t) - \cf_p(t)| }{\sqrt{\tf(t) (1 - \tf
(t))}} \leq 2\sqrt{2 p} \eta_p \bigl|\tf'(t)\bigr| /
\sqrt{\tf(t)} = 2\sqrt{2} p^{(1 -
\theta
)/2}\bigl |\tf'(t)\bigr| / \sqrt{\tf(t)}.
\]
Recall $s_p^* = \sqrt{2 \log(p)}$.
The above heuristic is captured in the following lemma, which is proved
in the
supplementary material \cite{HCTghsupp}.
%%%%%%%%%%%%%
%%%%%%%%%%%%%
%%%%%%%%%%%%%
%
%le3.3 #&#
\begin{lemma}\label{lemmaHCbias}
Fix $(\beta, r, \theta, a) \in(0,1)^4$. In the $\operatorname{ARW}(\beta, r,
\theta,
\Omega)$ model with $\Omega\in\mathcal{M}_p^*(a,K_p)$, there exists a
constant $C>0$ such that with probability at least $1-o(1/p)$, for all
$t$ such that $\bpsi^{-1}(\frac{1}{2}) < t < s_p^*$,
\[
\sqrt{p} \bigl|\cf_p(t) - \tf_p(t)\bigr| \cdot\bigl[\tf(t)
\bigl(1-\tf(t)\bigr)\bigr]^{-1/2} \leq L_p \max \bigl\{ \bigl(
p^{(1 - \theta)} \tf(t) \bigr)^{1/2}, 1\bigr\}.
\]
\end{lemma}
Combining Lemmas \ref{lemmastochunif} and \ref{lemmaHCbias}, we have
the following theorem, which is proved in the supplementary material
\cite{HCTghsupp}.
%%%%%%%%%%%%
%%%%%%%%%%%%
%%%%%%%%%%%%
%
%th3.1 #&#
\begin{thmm}\label{thmmHCfunction}
Fix $(\beta, r, \theta, a) \in(0,1)^4$. In the $\operatorname{ARW}(\beta, r,
\theta,
\Omega)$ model with $\Omega\in\mathcal{M}_p^*(a,K_p)$,
as $p \goto\infty$, with probability at least $1-o(p^{-1})$,
\[
\bigl|\operatorname{HC}(t, \cf_p) - \operatorname{HC}(t, \tf) \bigr| \leq L_p \bigl[
\bigl(p^{1 - \theta} \tf(t)\bigr)^{1/2} +1\bigr]\qquad \forall\bpsi
^{-1}\bigl(\tfrac{1}{2}\bigr) < t < s_p^*.
\]
\end{thmm}
By Theorem~\ref{thmmHCfunction}, in order for $|T_{\mathrm{HC}}(\cf_p) -
T_{\mathrm{HC}}(\tf)|$ to be small, we must have that for all $t$ in the
vicinity of $T_{\mathrm{HC}}(\tf)$,
%%%\[ L_p[(p^{1 - \theta} \tf(t))^{1/2} +1] \ll HC(t, \tf). \]
%
\[
\bigl|\operatorname{HC}(t,\cf_p) - \operatorname{HC}(t, \tf) \bigr| \ll \operatorname{HC}(t, \tf).
\]
When $\theta> 1/2$, this holds for all $(\beta, r)$ in Region of
Possibility since it can be checked that $L_p [(p^{1 - \theta} \tf
(t))^{1/2} +1] \ll \operatorname{HC}(t,\tf)$. When $\theta\leq1/2$,
this might not hold for all $(\beta, r)$ in this region, as the
estimation error of $\hat{\Omega}$ is simply too large. This explains
why we need to restrict HCT to be no less than $\tilde{s}_{p,n}^*$ as
in (\ref{Definesnp}). This also explains why we need conditions
(a)--(b) in
Theorem~\ref{thmmUB1}, but we do not need such conditions in Theorem~\ref{thmmIdealized} and Corollary~\ref{corUB}.

In the $\operatorname{ARW}(\beta, r, \theta, \Omega)$ model, $n_p = p^{\theta}$. Therefore,
\[
\tilde{s}_{p,n}^* = s_p(\theta)\qquad \mbox{if we let
$s_p(\theta) = \sqrt{2 \max\bigl\{(1 - 2 \theta), 0\bigr\}
\log(p)}$};
\]
see (\ref{Definesnp}). Accordingly, the HCT defined in (\ref
{HCTrefined}) can be rewritten as
\[
t_p^{\mathrm{HC}} = \cases{ %
T_{\mathrm{HC}}(\cf_p), &\quad $\mbox{if } s_p(\theta) \leq
T_{\mathrm{HC}}(\cf_p) \leq s_p^*$,
\vspace*{2pt}\cr
s_p(\theta), &\quad $\mbox{if } T_{\mathrm{HC}}(\cf_p) <
s_p(\theta)$,
\vspace*{2pt}\cr
s_p^*, &\quad $\mbox{if } T_{\mathrm{HC}}(\cf_p) >
s_p^*$. }
\]
It is worthy to note here that the ideal threshold always falls below
$s_p^*$, which is defined as $\sqrt{2 \log(p)}$; see Section~\ref{subsecidealtadd} and especially (\ref{eqdeftp^}).
It is also worthy to note that when $\theta< 1/2$ and when $\Omega$ is
unknown, the estimation error of $\Omega$ may have a major effect over
the classification
error,
especially when the threshold is small.
To alleviate such an effect, one possible approach is to set a number
$s_p(\beta, r, \theta)$ (say), and never allow the threshold to be
smaller than
$s_p(\beta, r, \theta)$. Since $(\beta, r)$ are unknown to us [but
$\theta\equiv\log(n_p)/\log(p)$
is known to us], so from a practical perspective, we must select
$s_p(\beta, r, \theta)$ in a way so that
it does not depend on $(\beta, r)$. Our calculations show that
$s_p(\theta) = \sqrt{2 \max\{(1 - 2 \theta), 0\} \log(p)}$ is one of
such choices.

The main result in this section is as follows, which is proved in
the supplementary material \cite{HCTghsupp}.
%%%%%%%%%%%%%%%
%%%%%%%%%%%%%%%
%%%%%%%%%%%%%%%
%%%%%%%%%%%%%%%
%
%th3.2 #&#
\begin{thmm}\label{thmmHCT}
Fix $(\beta, r, \theta, a) \in(0, 1)^4$ such that $(1 - \theta)/2 <
\beta<1-\theta$ and $r > \rho_{\theta}^*(\beta)$. In the
$\operatorname{ARW}(\beta,
r, \theta, \Omega)$ model with $\Omega\in\mathcal{M}_p^*(a,K_p)$,
\begin{longlist}[(1)]
\item[(1)] If $\theta> \frac{1}{2}$, then as $p \goto\infty$, there
are positive constants $c_2=c_2(\beta,r,a, \theta)$ and $d_0 =
d_0(\beta
, r, a, \theta)$ such that with probability at least $1-o(1/p)$,
$|t_p^{\mathrm{HC}} - T_{\mathrm{ideal}}(\eps_p, \tau_p, \Omega)|\leq L_pp^{-c_2}$ when
$r < \beta$, and
$t_p^{\mathrm{HC}} \in[\sqrt{2\beta\log p}- \Delta_1,\tau_p)$ when $r \geq
\beta
$, where $\Delta_1 = d_0 \log(\log(p))/\sqrt{\log(p)}$.

\item[(2)] If $0 <\theta\leq\frac{1}{2}$ and $(\beta, r, \theta)$
satisfy the conditions in Theorem~\ref{thmmUB1},
then with probability at least $1-o(1/p)$,
$|t_p^{\mathrm{HC}} - T_{\mathrm{ideal}}(\eps_p, \tau_p, \Omega)|\leq L_pp^{-c_3}$ for
some constant $c_3=c_3(\beta,r,a) > 0$ when $r < \beta$, and
$t_p^{\mathrm{HC}} \in[\sqrt{2\beta\log p}-\Delta_1,\tau_p)$ for $\Delta_1 =
d_1\log(\log( p))/\sqrt{\log p}$ when $r \geq\beta$, where $d_1 =
d_1(\beta,r, a) > 0$ is a constant.
\end{longlist}
\end{thmm}

%%%%%%%%%%%%%%%
%%%%%%%%%%%%%%%
%%%%%%%%%%%%%%%
%s3.2 #&#
\subsection{Stochastic fluctuation of Fisher's separation}
Similarly, the stochastic fluctuation of $\operatorname{Sep}(t, \hat{Z}, \mu, \hat
{\Omega})$ contains two parts:
that from $\tilde{Z} = \Omega Z$, and that from the estimation $\hat
{\Omega}$.
In detail,
\[
\bigl|\operatorname{Sep}(t, \hat{Z}, \mu, \hat{\Omega}) - \widetilde{\operatorname{Sep}}(t, \eps_p,
\tau _p, \Omega)\bigr| \leq2\cdot(I + \mathit{II}),
\]
where $I = \frac{1}{2} |\operatorname{Sep}(t, \tilde{Z}, \mu, \Omega) - \widetilde
{\operatorname{Sep}}(t, \eps_p, \tau_p, \Omega)|$ and $\mathit{II} = \frac{1}{2} |\operatorname{Sep}(t,
\hat
{Z}, \mu, \hat{\Omega}) - \operatorname{Sep}(t, \tilde{Z}, \mu, \Omega)|$.

Consider $I$ first. Recall that
\[
\operatorname{Sep}(t, \tilde{Z}, \mu, \Omega) = 2 M_p(t, \tilde{Z}, \mu, \Omega )
/ \sqrt{V_p(t, \tilde{Z}, \Omega)}.
\]
Heuristically,
$M_p(t, \tilde{Z}, \mu, \Omega) = m_p(t, \eps_p, \tau_p, \Omega) +
O_p(\sqrt{m_p(t, \eps_p, \tau_p, \Omega)})$ and\break $V_p(t, \tilde{Z},
\mu,
\Omega) = v_p(t, \eps_p, \tau_p, \Omega) + O_p(\sqrt{v_p(t, \eps
_p, \tau
_p, \Omega)})$; see (\ref{Definesmallmv}). Combining these with the
definitions, we expect that
%
%e3.2 #&#
%
\begin{eqnarray}
\label{mvapprox} &&\operatorname{Sep}(t, \tilde{Z}, \mu, \Omega)
\nonumber\\
&&\qquad= \widetilde{\operatorname{Sep}}(t,
\eps_p, \tau_p, \Omega)\\
&&\quad\qquad{}\times \biggl[1 + O_p \biggl(
\frac{1}{\sqrt{m_p(t, \eps_p, \tau_p,
\Omega)}} + \frac{1}{\sqrt{v_p(t, \eps_p, \tau_p, \Omega)}} \biggr)
\biggr],\nonumber
\end{eqnarray}
where in the square brackets, the second term is much smaller than $1$.
This is elaborated in the following lemma which is proved in the
supplementary material~\cite{HCTghsupp}. In detail, let
\[
q(t) = q(t; \beta, r, \theta, \Omega_p, p) = \cases{ %
p^{(1-\theta)/2 - \max\{4\beta-2r, 3\beta+r\}/4}, &\quad $r < \beta$,
\vspace*{2pt}\cr
0, &\quad $r \geq\beta.$}
\]

%%%%%%%%%%%%%%%%
%%%%%%%%%%%%%%%%
%%%%%%%%%%%%%%%%
%%%%%%%%%%%%%%%%
%
%le3.4 #&#
\begin{lemma} \label{lemmasepstoch}
Fix $(\beta, r, \theta, a) \in(0,1)^4$ such that $r > \rho_{\theta
}^*(\beta)$ and $(1-\theta)/2 < \beta< (1-\theta)$. In the
$\operatorname{ARW}(\beta,
r, \theta, \Omega)$ model with $\Omega\in\widetilde{\cal
M}^*_p(a,b,K_p)$, as $p \goto\infty$, with probability at least $1 - o(1/p)$,
\[
\sup_{\{ t > 0 \}} \bigl| \operatorname{Sep}(t, \tilde{Z}, \mu, \Omega) - \widetilde
{\operatorname{Sep}}(t, \eps_p, \tau_p, \Omega)\bigr| \leq L_p
\bigl[q(t) + p^{-\theta/2}\bigr].
\]
When $r < \beta$, the condition on $\Omega$ can be relaxed to that of
$\Omega\in{\cal M}_p^*(a, K_p)$.
\end{lemma}
Next, we consider $\mathit{II}$. The following lemma, which is proved in the
supplementary
material \cite{HCTghsupp}, characterizes the order of $\mathit{II}$.

%%%%%%%%%%%
%%%%%%%%%%%
%%%%%%%%%%%
%
%le3.5 #&#
\begin{lemma} \label{lemmasepbias}
Under the same conditions as in Lemma~\ref{lemmasepstoch}, as $p
\goto
\infty$,
with probability at least $1 - o(1/p)$, for all $t$ such that
$s_p(\theta)<t < s_p^*$,
$|\operatorname{Sep}(t, \hat{Z},\break \mu, \hat{\Omega}) - {\operatorname{Sep}}(t, \tilde{Z}, \mu,
\Omega
)| \leq L_p [p^{-\theta}(p\tf(t))^{1/2} + q(t) +p^{-\theta/2}]$.
When $r <\beta$, the condition on $\Omega$ can be relaxed to that of
$\Omega\in{\cal M}_p^*(a,K_p)$.
\end{lemma}
Combining Lemmas \ref{lemmasepstoch}--\ref{lemmasepbias} gives the
following theorem, the proof of which is omitted (note that Theorem~\ref
{thmmsepfunction} is parallel to Theorem~\ref{thmmHCfunction}).
%%%%%%%%%%%%%
%%%%%%%%%%%%%
%%%%%%%%%%%%%
%%%%%%%%%%%%%
%
%th3.3 #&#
\begin{thmm}\label{thmmsepfunction}
Under the same conditions as in Lemma~\ref{lemmasepstoch}, as $p
\goto
\infty$, with probability at least $1-o(p^{-1})$, for all $t$ such that
$s_p(\theta)<t <s_p^*$,
\[
\bigl|\operatorname{Sep}(t, \hat{Z}, \mu, \hat{\Omega}) - \widetilde{\operatorname{Sep}}(t, \eps_p,
\tau_p, \Omega)\bigr | \leq L_p \bigl[p^{-\theta}\bigl(p
\tf(t)\bigr)^{1/2}+ q(t) + p^{-\theta/2}\bigr].
\]
When $r <\beta$, the condition on $\Omega$ can be relaxed to that of
$\Omega\in{\cal M}_p^*(a,K_p)$.
\end{thmm}
%
%%%%%%%%%%%%%%
%%%%%%%%%%%%%%
%%%%%%%%%%%%%%
%s3.3 #&#
\subsection{\texorpdfstring{Proof of Theorems \protect\ref{thmmIdealized}--\protect\ref{thmmUB1}}
{Proof of Theorems 1.2--1.3}}
\label{subsecclassify}
We are now ready to prove Theorems \ref{thmmIdealized}--\ref{thmmUB1},
where $\Omega$ is assumed as known and unknown, respectively.
The proofs are similar, so we only show Theorem~\ref{thmmUB1}.
Consider $L_{\mathrm{HC}}(X, \hat{\Omega})$, where
$\hat{\Omega}$ is an acceptable estimator. The misclassification
error is
%
%e3.3 #&#
%
\begin{equation}
\label{eqclasserr1} P \bigl( Y \cdot L_{\mathrm{HC}}(X,\hat{\Omega}) < 0 \bigr)
= E_{\eps_p,
\tau
_p} E \bigl[ \bphi \bigl( \tfrac{1}{2}\operatorname{Sep}
\bigl(t_p^{\mathrm{HC}}, \hat{Z}, \mu, \hat {\Omega}\bigr) \bigr)
\bigr].
\end{equation}
We now prove for the case of $r < \beta$ and $r \geq\beta$ separately.

In the first case,\vspace*{1pt} we note that
$L_p[p^{-\theta}(p\tf(t))^{1/2}+ p^{-\theta/2}]\leq L_pp^{\min\{0,
{1}/{2}-\theta\}}$ for $ s_p(\theta)<t<s_p^*$. Write $T_{\mathrm{ideal}} =
T_{\mathrm{ideal}}(\eps_p, \tau_p, \Omega)$ and
$\widetilde{\operatorname{Sep}}(t) = \widetilde{\operatorname{Sep}}(t, \eps_p, \tau_p, \Omega)$ for
short as before. By Theorem~\ref{thmmsepfunction}, with probability $1
- o(1/p)$,
%
%e3.4 #&#
%
\begin{eqnarray}
\label{eqclasserr2} &&\bigl|\operatorname{Sep}\bigl(t_p^{\mathrm{HC}}, \hat Z, \mu, \hat{
\Omega} \bigr) - \widetilde {\operatorname{Sep}}\bigl(t_p^{\mathrm{HC}}\bigr)\bigr|
\nonumber
\\[-8pt]
\\[-8pt]
\nonumber
&&\qquad \leq
L_p \bigl[ p^{\min\{0, {1}/{2}-\theta\}} + p^{{(1-\theta)}/{2} - \max\{\beta-{r}/{2}, {(3\beta+r)}/{4}\}}\bigr].
\end{eqnarray}
At the same time, by Theorem~\ref{thmmHCT}, with probability $1 - o(1/p)$,
$|t_p^{\mathrm{HC}} - T_{\mathrm{ideal}}|$ is algebraically small.
Note that $\widetilde{\operatorname{Sep}}(t)$ is a nonstochastic function, and that
in the vicinity of $T_{\mathrm{ideal}}$, the second derivative of $\widetilde
{\operatorname{Sep}}$ at $t$ has
the same magnitude as that of $\widetilde{\operatorname{Sep}}(t)$, up to a
multi-$\log
(p)$ term (the first derivative is $0$ at $t = T_{\mathrm{ideal}}$).
By Taylor's expansion and Lemma~\ref{lemmaidealsep},
%
%e3.5 #&#
%
\begin{equation}
\label{eqclasserr3} \widetilde{\operatorname{Sep}} \bigl(t_p^{\mathrm{HC}}\bigr) =
\bigl(1 + o(1)\bigr) \widetilde{\operatorname{Sep}} (T_{\mathrm{ideal}}) = L_pp^{{(1-\theta)}/{2}-\delta(\beta,r)},
\end{equation}
where $\delta(\beta, r)$ is as in (\ref{eqdefdelta}).
By definitions, $\max\{4\beta-2r, 3\beta+r\}/4 > \delta(\beta,r)$.
Inserting (\ref{eqclasserr2})--(\ref{eqclasserr3}) into (\ref
{eqclasserr1}) gives
%
%e3.6 #&#
%
\begin{equation}
\label{classification11}\quad P \bigl( Y \cdot L_{\mathrm{HC}}(X, \hat{\Omega}) < 0
\bigr) = \bigl(1 + o(1/p) \bigr)\bphi \bigl(L_p p^{{(1-\theta)}/{2}-\delta(\beta
,r)}
\bigr) + o(1/p),
\end{equation}
and the claim follows since $(1 - \theta)/2 - \delta(\beta, r) > 0$.

In the second case, $\sqrt{2\beta\log p} \lesssim t_p^{\mathrm{HC}} \lesssim
\sqrt{2r\log p}$ with probability at least $1 - o(1/p)$. Combining
this with
Theorem~\ref{thmmsepfunction}, with probability at least $1 - o(1/p)$,
%
%e3.7 #&#
%
\begin{equation}
\label{eqclasserr2add}\bigl |\operatorname{Sep}\bigl(t_p^{\mathrm{HC}}, \hat Z, \mu,
\hat{\Omega} \bigr) - \widetilde {\operatorname{Sep}}\bigl(t_p^{\mathrm{HC}}\bigr)\bigr|
\leq L_p p^{\min\{0, {1}/{2}-\theta\}}.
\end{equation}
At the same time, by similar argument as that of the proof of Theorem~\ref{thmmIdealHCTb},
\[
2\tau_pK^{-1}_pp^{(1-\theta-\beta)/2}\lesssim
\widetilde {\operatorname{Sep}}\bigl(t_p^{\mathrm{HC}}\bigr) \leq
\widetilde{\operatorname{Sep}}(T_{\mathrm{ideal}}) = L_pp^{(1-\theta
-\beta)/2}.
\]
Combining this with (\ref{eqclasserr1}) and (\ref{eqclasserr2add}) gives
%
%e3.8 #&#
%
\begin{equation}
\label{classerr2} \qquad P \bigl( Y \cdot L_{\mathrm{HC}}(X, \hat{\Omega}) < 0 \bigr) =
\bigl(1+o(1/p)\bigr) \bphi \bigl( \tfrac{1}{2}L_pp^{(1-\theta)/2 -\delta(\beta,r) }
\bigr) + o(1/p),
\end{equation}
and the claim follows since $\frac{1 - \theta}{2} - \delta(\beta,
r) >
0$. This proves Theorem~\ref{thmmUB1}.

We conclude this section by a remark on the convergence rate.
At the end of Section~\ref{secIdealHC}, we show that
the ``ideal'' classifier $L_t(X, \Omega)$ has very fast convergence rate
with $t$ being either the ideal threshold or the ideal HCT. In
comparison, the convergence rate of $L_{\mathrm{HC}}(X, \hat{\Omega})$ is
unfortunately much slower (but is still algebraically fast). To explain
this, we note that the rate of convergence of $t_p^{\mathrm{HC}}$ to $T_{\mathrm{HC}}(\tf
)$ and the rate of convergence of $\hat{\Omega}$ to $\Omega$ are both
algebraically fast; if these convergence rates can be improved, then
the misclassification error rate of $L_{\mathrm{HC}}(X, \hat{\Omega})$ can be
improved as well.

%%%%%%%%%%%%%%%
%%%%%%%%%%%%%%%
%%%%%%%%%%%%%%%
%%%%%%%%%%%%%%%

%%%%%%%%%%%%%%
%%%%%%%%%%%%%%
%%%%%%%%%%%%%%
%%%%%%%%%%%%%%
%%%%%%%%%%%%%%
%%%%%%%%%%%%%%
%%%%%%%%%%%%%%
%%%%%%%%%%%%%%
%%%%%%%%%%%%%%
%%%%%%%%%%%%%%
%s4 #&#
\section{Simulations} \label{secSimul}
We have conducted a small-scale numerical study. The idea is to select
a few sets of representative parameters for experiments, and compare
the performance of HCT classifier (HCT) with three other methods:
ordinary HCT (oHCT),
pseudo HCT (pHCT), and CVT.
All these methods are very similar to HCT, except that (a) in pHCT, we
assume $\Omega$ is known to us, (b) in CVT, we set the threshold of IT
by a $5$-fold cross validation, and (c) in oHCT, we pretend $\Sigma$ is
diagonal, and estimate $\Omega$ accordingly.
Note that CVT reduces to PAM \cite{TibshiraniPNAS2002} if we do not utilize
the correlation structure; see more discussion in \cite{DonohoJin2008}.

%%%%%%%%%%%
%%%%%%%%%%%
%%%%%%%%%%%
%s4.1 #&#
\subsection{\texorpdfstring{Estimating $\Omega$}{Estimating Omega}}
For some of the procedures, we need to estimate $\Omega$.
We use Bickel and Levina's Thresholding (BLT) procedure \cite
{Bickel2008}. Alternatively, one could use the glasso \cite
{FriedmanHastieTibshirani2007} or the CLIME \cite{CaiLuo2011}. But since
the main goal is to investigate the performance of HCT, we do not
include glasso and CLIME in the study: if HCT performs well with
$\Omega
$ estimated by BLT,
we expect it to perform even better if $\Omega$ is estimated more accurately.

At the same time, each of these methods can be improved numerically
with an additional \textit{re-fitting} stage. Take the BLT for example.
For the training data $\{(X_i, Y_i)\}_{i = 1}^n$, let
$\bar{X} = \frac{1}{n} \sum_{i = 1}^n Y_i X_i$, and let
$\hat{\Sigma} = \frac{1}{n} \sum_{i = 1}^n (Y_i X_i - \bar{X})' (Y_i
X_i - \bar{X})$ be the
empirical covariance matrix. BLT starts by obtaining an estimate of
$\Sigma$ using thresholding:
%
%e4.1 #&#
%
\begin{equation}
\label{BL} \Sigma^*(i,j) = \hat{\Sigma}(i,j) 1\bigl\{\bigl|\hat{\Sigma}(i,j)\bigr| \geq
\eta \bigr\}, \qquad 1 \leq i, j \leq p,
\end{equation}
and then estimate $\Omega$ by $\hat{\Omega}^{**} = (\Sigma^*)^{-
1}$. Here,
$\eta> 0$ is a tuning parameter.\eject

We propose the following refitting stage to improve the estimator.
Fixing a tuning parameter $\zeta> 0$, we further improve $\hat{\Omega
}^{**}$ via coordinate-wise thresholding and call the resultant
estimator $\hat{\Omega}^*$:
%
%e4.2 #&#
%
\begin{equation}
\label{zeta} \hat{\Omega}^*(i,j) = \hat{\Omega}^{**}(i,j) 1\bigl\{ \bigl|
\hat{\Omega }^{**}(i,j)\bigr| \geq\zeta\bigr\}.
\end{equation}
For each $1 \leq i \leq p$, let $S_i = \{1 \leq j \leq p\dvtx \hat{\Omega
}^*(i,j) \neq0\}$, and let
$A_i$ be the sub-matrix of $\hat{\Sigma}$ formed by restricting the
rows/columns of $\hat{\Sigma}$ to $S_i$. Denote the final estimate of
$\Omega$ by $\hat{\Omega} = [\omega_1, \omega_2, \ldots, \omega
_p]$. We
define $\omega_i$ as follows.
Write $S_i = \{j_1, j_2, \ldots, j_k\}$, where $k = |S_i|$. Let $e_i$
be the $p \times1$ vector such that $e_i(j) = 1\{i = j\}$, $1 \leq j
\leq p$, and let $\xi_i$ be the $k \times1$ vector formed by
restricting the rows of $e_i$ to $S_i$. Define $\eta_i = A_i^{-1} \xi
_i$. We let $\omega_i(j_{\ell}) = \eta_i(\ell)$, $1 \leq\ell\leq k$,
and let $\omega_i(j) = 0$ if $j \notin S_i$.

The resultant estimation of the refitting procedure is a symmetric
matrix, which is also positive definite, provided
that $K_p$ is sufficiently small (say, $\sqrt{\log(p)} K_p \ll\sqrt {n}$) and that the smallest eigenvalue of $\Omega$ is bounded from
below by a constant $C > 0$; recall that $K_p$ is the maximum of the
number of nonzeros in each row of~$\Omega$.

%s4.2 #&#
\subsection{Numerical experiments} \label{numexp}
Fix $(p, n, \eps_p, H_p, \Omega)$ and an integer $m$, each simulation
experiment contains the following main steps.
\begin{longlist}[1.]
\item[1.] Generate a $p \times1$ vector $\mu$ according to $(\sqrt{n}
\mu
(j)) \stackrel{\mathrm{i.i.d.}}{\sim} (1-\varepsilon_{p}) \nu_{0}+\varepsilon_{p} H_{p}$.
\item[2.] Generate training data $(X_i, Y_i)$, $1 \leq i \leq n$, by
letting $Y_i = 1$ for $i \leq n/2$ and $Y_i = -1$ for $i > n/2$, and
$X_i \sim N(Y_i \cdot\mu, \Omega^{-1})$.
\item[3.] Generate $m$ test vectors, each of which has the form of $X \sim
N(Y \cdot\mu, \Omega^{-1})$, where $Y = \pm1$ with equal probabilities.
\item[4.] Use the training data to build all four classifiers (HCT, oHCT,
pHCT and CVT), apply them to the test set, and then record the test errors.
\end{longlist}
When we need to estimate $\Omega$, we use BLT with the aforementioned
refitting stage. The study contains three different experiments, which
we now discuss separately.

\textit{Experiment} 1. In this experiment, we compare HCT with oHCT and
pHCT. The experiment contains three sub-experiments 1a, 1b and 1c.

\begin{figure}

\includegraphics{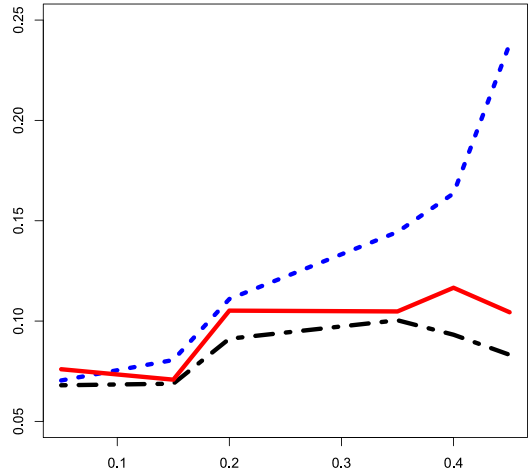}

\caption{Comparison of classification errors by HCT (solid), oHCT
(dashed) and pHCT (dash-dotted). The x-axis is $a$, and the y-axis is
the classification error (Experiment \textup{1a}).}
\label{figfig1}
\end{figure}

In Experiment 1a, we fix $(p, n, \eps_p, \tau_p, m)=(3000, 2000, 0.1,
4, 500)$, and let $H_p$ be the point mass at $\tau_p$. Also, we choose
$\Omega$ to be the tridiagonal matrix
%
%e4.3 #&#
%
\begin{equation}
\label{eqtridiag} \Omega(i,j) = 1 \{ i = j\} + a \cdot1\bigl\{|i - j|= 1\bigr\},\qquad 1 \leq i,
j \leq p,
\end{equation}
where $a$ takes values from $\{0.05, 0.15, 0.2, 0.35, 0.4, 0.45\}$. The
results are reported in Figure~\ref{figfig1}. The tuning parameter
$\eta$ in (\ref{BL}), which varies with the values of $a$, $n$ and $p$,
is calculated from trials of comparing $(\Sigma^*)^{- 1}$ with the true~$\Omega$.
The tuning parameter $\zeta$ in (\ref{zeta}), which also
varies with the values of $a$, $n$ and~$p$, is chosen so that there are
only $k$ nonzero coordinates in each row\vadjust{\goodbreak} of $\hat{\Omega}^*$ after
thresholding of $\hat{\Omega}^{**}$. We let $k=2,3$ if $\Omega$ is
tridiagonal and
$k=4,5$ if $\Omega$ is five-diagonal (see experiments below). In this
experiment, $\eta$ is set accordingly from $\{0.1, 0.1, 0.15, 0.15, 0.2,
0.25\}$ and $\zeta$ is from $\{0.05, 0.1, 0.1, 0.2, 0.25, 0.3\}$. The results
suggest that HCT outperforms oHCT, but is slightly inferior to pHCT
since we have to pay a price for estimating $\Omega$. As $a$ increases,
the correlation structure becomes increasingly influential, so
the advantage of HCT over oHCT becomes increasingly prominent (but
differences between
HCT and pHCT remain almost the same).

%%%%%%%%%%
%%%%%%%%%%
%%%%%%%%%%
%%%%%%%%%%

%%%%%%%%%%%%%%%%
%%%%%%%%%%%%%%%%
%%%%%%%%%%%%%%%%
%
%f1 #&#

%%%%%%%%%%

In Experiment 1b, for various $(p, n, \eps_p, \tau_p)$, we choose
$m=500$ and let $\Omega$ be either of the following tridiagonal matrix
or five-diagonal matrix. In the first case, $\Omega$ is a $p \times p$
tridiagonal matrix with $1$ on the diagonal and $a$ on the
off-diagonal. In the second case, $\Omega$ is a $p \times p$
five-diagonal matrix with $1$ on the diagonal, $a_1$ on the first
off-diagonal, and $a_2$ on the second off-diagonal. Experiment 1c uses
a very similar setting, except that we take $H_p$ as the uniform
distribution over $[\tau_{p}-0.5, \tau_{p}+0.5]$. We select $\zeta$ and
$\eta$ similarly as in Experiment 1a. The results based on 25
repetitions for Experiment 1b--1c are reported in Table~\ref{tabletab1}, which suggest that HCT outperforms oHCT and that pHCT
slightly outperforms HCT.

%%%%%%%%%%
%%%%%%%%%%
%%%%%%%%%%
%
%t1 #&#
\begin{table}
\caption{Classification errors by HCT, oHCT and pHCT. $\Omega$ is
tridiagonal (left two columns) or five-diagonal matrix (right column).
Rows 1--2: Experiment \textup{1b}. Row 3: Experiment \textup{1c}}\label{tabletab1}
\begin{tabular*}{\textwidth}{@{\extracolsep{\fill}}lccc@{}}
% used for centering table
%
\hline%inserts double horizontal lines
& \multicolumn{1}{c}{$\bolds{n=1000, p=2000}$} &
\multicolumn{1}{c}{$\bolds{n=2000, p=3000}$} &
\multicolumn{1}{c}{$\bolds{n=2000}$\textbf{,} $\bolds{p=3000}$} \\
%inserts table
& \multicolumn{1}{c}{$\bolds{a=0.05, \varepsilon_{p}=0.1} $}&
\multicolumn{1}{c}{$\bolds{a=0.45}$\textbf{,} $\bolds{\varepsilon_{p}=0.2}$}&
\multicolumn{1}{c@{}}{$\bolds{a_1=0.45}$\textbf{,} $\bolds{a_2 =0.2}$} \\
& \multicolumn{1}{c}{$\bolds{\tau_{p}=4}$} &
\multicolumn{1}{c}{$\bolds{\tau_{p}= 3}$}&
\multicolumn{1}{c@{}}{$\bolds{\varepsilon_{p}=0.1, \tau_{p}=4}$} \\
 % inserts table
\hline% inserts single horizontal line
oHCT & 0.0508 & 0.2818 & 0.1492 \\ % inserting body of the table
pHCT & 0.0384 & 0.0698 & 0.1015 \\
HCT & 0.0523 & 0.0742 & 0.1053 \\
\hline\\[3pt]
%inserts double horizontal lines
& \multicolumn{1}{c}{$\bolds{n=500, p=1000}$} &
\multicolumn{1}{c}{$\bolds{n=2000}$\textbf{,} $\bolds{p=3000}$}&
\multicolumn{1}{c}{$\bolds{n=2000}$\textbf{,} $\bolds{p=3000}$} \\
%inserts table
& \multicolumn{1}{c}{$\bolds{a=0.05, \varepsilon_{p}=0.1}$} &
\multicolumn{1}{c}{$\bolds{a = 0.45}$\textbf{,} $\bolds{\varepsilon_{p}=0.05}$} &
\multicolumn{1}{c@{}}{$\bolds{a_1 = 0.35}$\textbf{,} $\bolds{a_2 = 0.2}$\textbf{,}} \\
& \multicolumn{1}{c}{$\bolds{\tau_{p}=4}$} &
\multicolumn{1}{c}{$\bolds{\tau_{p}=5}$} &
\multicolumn{1}{c@{}}{$\bolds{\varepsilon_{p}=0.05}$\textbf{,} $\bolds{\tau_{p}=4}$} \\
\hline% inserts single horizontal line
oHCT & 0.0560 & 0.2629 & 0.2183 \\ % inserting body of the table
pHCT & 0.0571 & 0.1398 & 0.1893 \\
HCT & 0.0572 & 0.1438 & 0.1959 \\
\hline\\[3pt]
&  &  &\multicolumn{1}{c@{}}{$\bolds{n=2000}$\textbf{,} $\bolds{p=3000}$} \\
& \multicolumn{1}{c}{$\bolds{n=1000, p=2000}$} &
\multicolumn{1}{c}{$\bolds{n=2000, p=3000}$} &
$\bolds{H_{p}=U(3.5,4.5)}$ \\
& $\bolds{H_{p}=U(3.5,4.5)}$ &
$\bolds{H_{p}=U(2.5,3.5)}$ &
$\bolds{a_1=0.45, a_2=0.2}$ \\ % inserts table
& $\bolds{a=0.05, \varepsilon_{p}=0.1}$ &
$\bolds{a=0.45, \varepsilon_{p}=0.2}$ &
$\bolds{\varepsilon_{p}=0.1}$ \\ % inserts table
\hline% inserts single horizontal line
oHCT & 0.0444 & 0.2672 & 0.1648 \\ % inserting body of the table
pHCT & 0.0522 & 0.0733 & 0.1159 \\
HCT & 0.0508 & 0.0843 & 0.0977 \\
\hline %inserts single line
\end{tabular*}
\end{table}

\textit{Experiment} 2. In this experiment, we compare the pHCT with the
CVT assuming $\Omega$ is known (the case $\Omega$ is unknown is
discussed in Experiment 3). Experiment~2 contains two sub-experiments,
2a and 2b.

In Experiment 2a, we consider $6$ different combinations of $(p, n,
\eps
_p, \tau_p)$ with $m=500$, and let $\Omega$ be the tridiagonal matrix
as in (\ref{eqtridiag}) with $a = 0.2$. Averages
of the selected thresholds and classification errors
across different replications are reported in Table~2. The results over
25 repetitions suggest that
the threshold choices by HC and cross validations
are considerably different, with the former being more accurate and
more stable.
Note that HCT is also computationally much more efficient than the CVT.

%%%%%%%%%%%%
%%%%%%%%%%%%
%%%%%%%%%%%%
%
%t2 #&#
\begin{table}[b]
\caption{Comparison of thresholds (Column 2, 4, 6) and classification
errors (Column 3, 5, 7) by pHCT and CVT. $(p, \tau_p) = (3000, 1.8)$,
and $\eps_p = 0.1$ (top) and $0.05$ (bottom). Left to right: $n = 100,
50, 20$ (Experiment \textup{2a})}\label{tabletab2}
% used for centering table
%
\begin{tabular*}{\textwidth}{@{\extracolsep{\fill}}ld{1.2}d{1.3}d{1.2}d{1.3}d{1.2}d{1.2}@{}}
\hline%inserts double horizontal lines
& \multicolumn{1}{c}{\textbf{Threshold}} & \multicolumn{1}{c}{\textbf{Error}} &\multicolumn{1}{c}{\textbf{Threshold}} &
\multicolumn{1}{c}{\textbf{Error}}& \multicolumn{1}{c}{\textbf{Threshold}} & \multicolumn{1}{c@{}}{\textbf{Error}}\\
%% inserts table
\hline% inserts single horizontal line
pHCT & 1.9 & 0.05 & 2.16 & 0.002 & 1.99 &0  \\ % inserting body of the
%table
CVT & 2.5 & 0.08 & 1 & 0.018 & 1 &0 \\[3pt]
%inserts double horizontal lines
pHCT & 2.39 & 0.18 & 2.06 & 0.10 & 2.13 &0.02  \\ % inserting body of
%the table
CVT & 1.9 & 0.224 & 2.00 & 0.14 & 1.1 &0.09\\
\hline%inserts single line
\end{tabular*}
\end{table}

\begin{figure}

\includegraphics{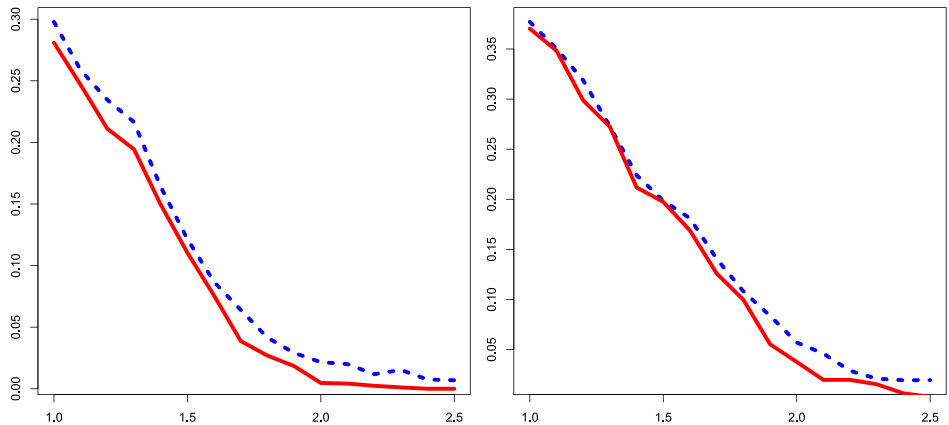}

\caption{Classification errors of pHCT (solid) and CVT (dashed) for $n
= 20$ (left) and $40$ (right) and various $\tau_p$ (x-axis)
(Experiment \textup{2b}).}
\label{figfig2} % \caption{}\label{}
\end{figure}

In Experiment 2b, we set $(p, \eps_p, m) =(3000, 0.05, 500)$, $n \in\{
20, 40\}$, and let $\Omega$ be the same as in Experiment 2a. We let
$\tau_p$ range from $1$ to $2.5$ with an increment of $0.1$. The
classification errors over 25 repetitions by pHCT and CVT are in Figure~\ref{figfig2}, where a conclusion similar to that in Experiment 2a
can be drawn.

%%%%%%%%%%%%%
%%%%%%%%%%%%%
%%%%%%%%%%%%%
%%%%%%%%%%%%%
%
%f2 #&#

\begin{figure}

\includegraphics{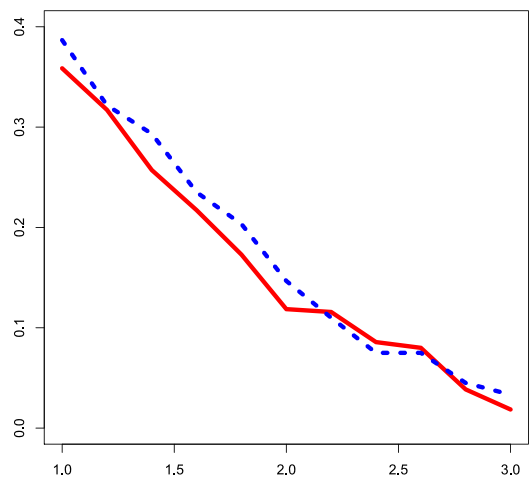}

\caption{Classification errors by HCT (solid) and CVT (dashed) for
various $\tau_{p}$ (x-axis) (Experiment~3).}
\label{figfig3}
\end{figure}

\textit{Experiment} 3. We compare the performance of HCT with CVT for the
case where $\Omega$ is unknown and needs to be estimated. Note that for
small $n$ (say, less than 500) we might not have reasonable accuracy on
estimating $\Omega$ using BLT. For small $p$, say $100$--$300$, the CVT
is computationally very slow and it is very likely that the refitting
procedure for BLT would not have decent performance. We take $(p, n,
\varepsilon_{p})=(5000, 500, 0.1)$ and let $\Omega$ be the block diagonal
matrix consisting $10$ diagonal blocks, each is a big five-diagonal
matrix $C = C_{500,500}(a_1, a_2)$, where $C(i,j) = 1\{i = j\} + a_1
\cdot1\{|i - j| = 1\} + a_2 \cdot1\{|i - j| = 2\}$, $1 \leq i, j
\leq500$, and $a_1=0.45$, $a_2=0.1$. We let $\tau_p$ range from $1$ to
$3$ with an increment of 0.2. The tuning parameters $\zeta$ and $\eta$
are set in the same way as in Experiment 1. The results are reported in
Figure~\ref{figfig3}. Due to high computational cost, we only conduct
$m = 6$ repetitions, so the results are a bit noisy. Still, it is seen
that HCT outperforms CVT.

%%%%%%%%%%%%%%%%%%%%
%%%%%%%%%%%%%%%%%%%%
%%%%%%%%%%%%%%%%%%%%
%
%f3 #&#

In summary, for a reasonably large sample size $n$, HCT outperforms
oHCT and is only slightly inferior to pHCT. The reason we need a
relatively large $n$ is mainly due to that
we need to estimate $\Omega$. The relative performance of pHCT, HCT,
and oHCT is intuitive, since pHCT utilizes the true correlation
structure among the features, HCT estimates the correlation structure,
while oHCT ignores it. The comparisons of pHCT with CVT in Experiments
2a--2b suggest that if $\Omega$ is known, then HCT dominates CVT.
Experiment 3 shows that when $p$ is several times larger than $n$
(e.g., 10 times larger), HCT has smaller classification errors than CVT
does, and the precision matrix $\Omega$ can bez estimated reasonably well.

For larger $p$, the advantages of the HCT are even more prominent than
those considered here. We skip the comparisons for larger $p$ due to
high computational cost, which mainly comes from the BLT procedure (we must
run the algorithm many times to select a good tuning parameter $\eta$).
In the future, if we could find a more efficient method for estimating
$\Omega$, then HCT will be both more effective and more convenient to
use for large $p$.


\begin{supplement}[id=suppA]
\stitle{Supplementary material for ``Optimal classification in sparse
Gaussian graphic model''}
\slink[doi]{10.1214/13-AOS1163SUPP} %[doi,text={...}] - jei reikia
%suskaldyti doi
\sdatatype{.pdf}
\sfilename{aos1163\_supp.pdf}
\sdescription{We include all technical proofs omitted from the main text.}
\end{supplement}

%
% imsref loaded by akundreckaite, 2013-09-23 09:47:08
%

%

\printaddresses

\end{document}